\documentclass[runningheads]{llncs}
\usepackage{graphicx}
\usepackage{tikz}
\usepackage{comment}
\usepackage{amsmath,amssymb} % define this before the line numbering.
\usepackage{color}
\usepackage{orcidlink}

\usepackage[accsupp]{axessibility}  % Improves PDF readability for those with disabilities.
\makeatletter
\def\blfootnote{\xdef\@thefnmark{}\@footnotetext}
\makeatother

\usepackage[caption=false]{subfig}
\usepackage{caption}
\usepackage{booktabs}
\usepackage{multirow}
\usepackage{makecell}
\usepackage{algorithmic}
\usepackage{balance}
\usepackage{marvosym}
\usepackage{ifsym}
\begin{document}
% \linenumbers
\pagestyle{headings}
\mainmatter
\def\ECCVSubNumber{0016}  % Insert your submission number here

\title{Generative Domain Adaptation for Face Anti-Spoofing} % Replace with your title

\begin{comment}
\titlerunning{ECCV-22 submission ID \ECCVSubNumber} 
\authorrunning{ECCV-22 submission ID \ECCVSubNumber} 
\author{Anonymous ECCV submission}
\institute{Paper ID \ECCVSubNumber}
\end{comment}
%******************

%\begin{comment}
\titlerunning{Generative Domain Adaptation for Face Anti-Spoofing}
% \author{First E. van Author\inst{1}\orcidlink{0000-1111-2222-3333}\index{van Author, First E.} \and
% Second Author\inst{2,3}\orcidlink{1111-2222-3333-4444} \and
% Third Author\inst{3}\orcidlink{2222--3333-4444-5555}}

% \author{Qianyu Zhou\inst{1}$^*$$^\dagger$\orcidlink{0000-0002-5331-050X}\index{Qianyu, Zhou}  \and
% Ke-Yue Zhang\inst{2}$^*$\orcidlink{0000-0003-3589-5580}  \and
% Taiping Yao\inst{2}\orcidlink{0000-0002-2359-1523} \and
% Ran Yi\inst{1}\orcidlink{0000-0003-1858-3358}  \and \\
% Kekai Sheng\inst{2}\orcidlink{0000-0002-5382-3241}  \and
% Shouhong Ding\inst{2}$^\ddagger$ \orcidlink{0000-0002-3175-3553}  \and
% Lizhuang Ma\inst{1}$^\ddagger$ \orcidlink{0000-0003-1653-4341} 
% }
%
% 
\author{Qianyu Zhou\inst{1}$^*$$^\dagger$\orcidlink{0000-0002-5331-050X}\index{Qianyu, Zhou}  \and
Ke-Yue Zhang\inst{2}$^*$\orcidlink{0000-0003-3589-5580}\and
Taiping Yao\inst{2}\orcidlink{0000-0002-2359-1523} \and
Ran Yi\inst{1}\orcidlink{0000-0003-1858-3358}  \and \\
Kekai Sheng\inst{2}\orcidlink{0000-0002-5382-3241}  \and
Shouhong Ding\inst{2}\textsuperscript{\Letter}\orcidlink{0000-0002-3175-3553}  \and
Lizhuang Ma\inst{1}\textsuperscript{\Letter}\orcidlink{0000-0003-1653-4341} 
}

\authorrunning{Q. Zhou et al.}
\institute{Shanghai Jiao Tong University, Shanghai, China \and
Youtu Lab, Tencent, Shanghai, China\\
\email{\{zhouqianyu, ranyi\}@sjtu.edu.cn, ma-lz@cs.sjtu.edu.cn}\\
\email{\{zkyezhang,taipingyao,saulsheng,ericshding\}@tencent.com}}
%\end{comment}
%******************

    \maketitle
% $\blfootnote{* Equal contributions.}$ 
$\blfootnote{$*$ Equal contributions.}$ 
$\blfootnote{$\dagger$ Work done during an internship at Youtu Lab, Tencent.}$
$\blfootnote{\textsuperscript{\Letter} Corresponding authors.}$
% \textsuperscript{\rm 1}\footnotetext{\textsuperscript{\rm 1}
\begin{abstract}
Face anti-spoofing (FAS) approaches based on unsupervised domain adaption (UDA) have drawn growing attention due to promising performances for target scenarios. Most existing UDA FAS methods typically fit the trained models to the target domain via aligning the distribution of semantic high-level features. However, insufficient supervision of unlabeled target domains and neglect of low-level feature alignment degrade the performances of existing methods. To address these issues, we propose a novel perspective of UDA FAS that directly fits the target data to the models, \emph{i.e.,} stylizes the target data to the source-domain style via image translation, and further feeds the stylized data into the well-trained source model for classification. The proposed Generative Domain Adaptation (GDA) framework combines two carefully designed consistency constraints: 1) Inter-domain neural statistic consistency guides the generator in narrowing the inter-domain gap. 2) Dual-level semantic consistency ensures the semantic quality of stylized images. Besides, we propose intra-domain spectrum mixup to further expand target data distributions to ensure generalization and reduce the intra-domain gap. Extensive experiments and visualizations demonstrate the effectiveness of our method against the state-of-the-art methods.

\keywords{Face anti-spoofing, unsupervised domain adaptation}
\end{abstract}

\section{Introduction}
% \blfootnote{The first two authors contributed equally.~~Corresponding author: Lizhuang Ma and Shouhong Ding}
Face recognition (FR) techniques~\cite{deng2019arcface,kemelmacher2016megaface,taigman2014deepface,zhu2022local,li2021spherical,wang2021facex} have been widely utilized in identity authentication products, \emph{e.g.,} smartphones login, access control, \emph{etc}. 
Despite its gratifying progress in recent years, FR systems are vulnerable to face presentation attacks (PA), \emph{e.g.,} printed photos, video replay, and \emph{etc}. 
To protect such FR systems from various face presentation attacks, face anti-spoofing (FAS) attracts great attention. 
Nowadays, based on hand-crafted features~\cite{boulkenafet2015face,maatta2011face,LBP01,2014Context,HoG01}, and deeply-learned features~\cite{DeepBinary00,DeepBinary01,DeepBinary02,yu2021revisiting,zhang2021structure,lin2019face,disentangle01}, several methods achieve 
promising performance in intra-dataset scenarios. 
However, they all suffer from performance degradation when adapting to the target domains in real-world scenarios due to the domain gap across different domains.

\begin{figure}[t!]
	\centering
	\includegraphics[width=1.0\linewidth]{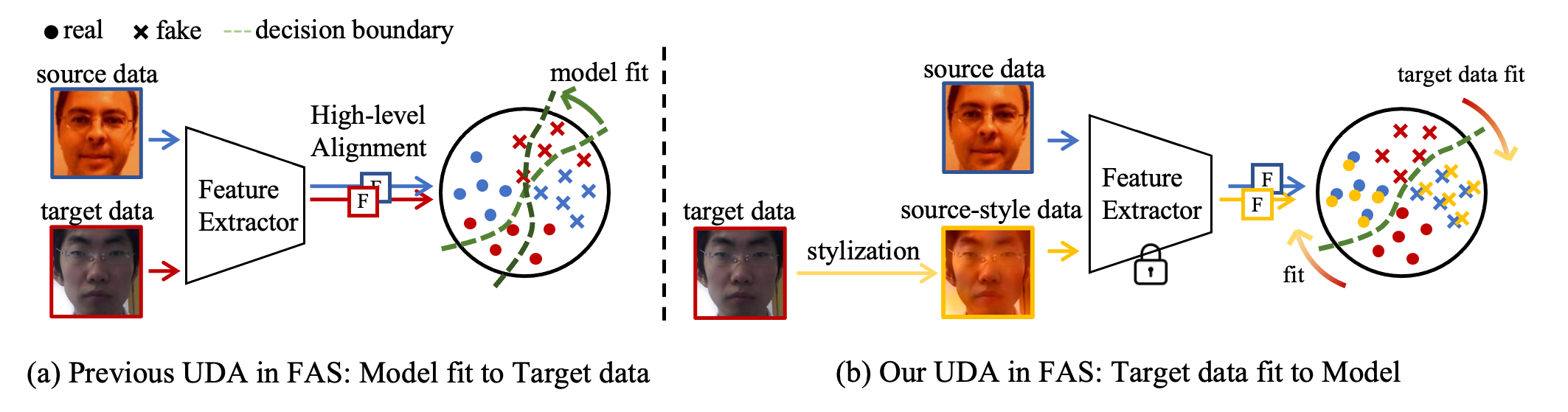}
	\caption{Conventional UDA FAS methods typically force the model fit to the target data via performing the high-level feature alignment across domains. 
	However, insufficient supervisions and neglect of low-level feature alignment inevitably affect the discrimination ability of FAS models. 
	Instead, we aim to directly fit the target data to the source-trained models in a reverse manner via
	both the high-level and low-level alignment.
	By generating source-style images and predicting with a well-trained model, we address these issues without changing models}
	\label{illustration}
\end{figure}

To improve the generalization, FAS approaches based on domain generalization (DG) and unsupervised domain adaption (UDA) have been proposed on cross-domain scenarios.
However, DG FAS approaches~\cite{2019Multi,2020Single,2020Regularized,liu2021dual,liu2021adaptive,chen2021generalizable,zhou2022adaptive}  only utilize the seen data in the training stage, which fail in utilizing the information of the target data, thus resulting in unsatisfactory performance on the target domain. 
Although UDA FAS methods~\cite{DR-UDA,2018Unsupervised,2019Improving,jia2021unified,quan2021progressive,zhou2019face,tu2019deep} utilize both the labeled source domain and the unlabeled target domain to bridge the domain gap, most of them typically fit the models to the target domain via aligning the distribution of semantic high-level features, as shown in Fig.~\ref{illustration} (a), without considering the specificity of FAS task. On the one hand, 
since the insufficient supervision of the target domain, fitting to it may inevitably affect the discrimination ability of the source model.
On the other hand, as pointed out in~\cite{2018De-Spoofing}, low-level features are especially vital to the FAS task. 
Thus, the above towards-target distribution alignment based on only high-level features may not be the most suitable way for UDA FAS.

To address the above issues, we propose a novel perspective of unsupervised domain adaptation (UDA) for face anti-spoofing (FAS). Different from existing methods that fit the models to the target data, we aim to directly fit the target data to the well-trained models, keeping the source-trained models unchanged, as shown in Fig.~\ref{illustration} (b).
To achieve such fitting, we reformulate the unsupervised domain adaptation (UDA) in FAS as a domain stylization problem to stylize the target data with the source-domain style, and the stylized data is further fed into the well-trained source model for classification.
In this work, we propose Generative Domain Adaptation (GDA) framework combining two carefully designed consistency constraints.
Specifically, we present inter-domain neural statistic consistency (NSC) to guide the generator toward producing the source-style images, which fully aligns the target feature statistics with the source ones in both high-levels and low-levels, %and is more suitable for the FAS task 
and narrows the inter-domain gap efficiently.
Besides, to maintain the semantic qualities and liveness information of the target data during the stylization procedure, we introduce a dual-level semantic consistency (DSC) on both image level and feature level. 
Moreover, intra-domain spectrum mixup (SpecMix) is presented to further expand the target data distribution to
ensure that the generator could correctly translate the unseen target domain to the source-style domain. To the best of our knowledge,
this is the first work that reveals the potential of image translation for UDA FAS.

Our main contributions can be summarized as follows:

$\bullet$ We propose a novel perspective of unsupervised domain adaptation for face anti-spoofing that directly fits the target data to the source model by stylizing the target data with the source-domain style via image translation.
    
$\bullet$ To ensure the stylization, 
we present a Generative Domain Adaptation framework 
combined with two carefully designed consistency constraints, inter-domain neural statistic consistency (NSC) and dual-level semantic consistency (DSC). 
And intra-domain spectrum mixup (SpecMix) is presented to further expand the target data distribution to ensure generalization.
    
$\bullet$ Extensive experiments and visualizations demonstrate the effectiveness of our proposed method against the state-of-the-art competitors.

\section{Related Work}
\subsubsection{Face Anti-Spoofing.}
% \noindent \textbf{Face Anti-Spoofing.} 
Face anti-spoofing (FAS) aims to detect a face image whether taken from a real person or various face presentation attacks~\cite{2017OULU,2012Replay,chen2022daptach,chen2022ecvit}. Pioneer works utilize handcrafted features to tackle this problem, such as SIFT~\cite{2016Secure}, LBP~\cite{boulkenafet2015face,maatta2011face,LBP01}, and HOG~\cite{2014Context,HoG01}. Several methods utilize the information from different domains, such as HSV and YCrCb color spaces~\cite{boulkenafet2015face,boulkenafet2016face}, temporal domains~\cite{siddiqui2016face,bao2009liveness}, and Fourier spectrum~\cite{li2004live}.
Recent approaches leverage CNN to model FAS with binary classification~\cite{DeepBinary00,DeepBinary01,DeepBinary02,zhang2021aurora} or 
additional supervision, \emph{e.g.,} depth map~\cite{yu2021revisiting}, reflection map~\cite{zhang2021structure} and r-ppg signal~\cite{lin2019face,hu2021end}. Other methods adopt disentanglement~\cite{disentangle01,STCN} and custom operators~\cite{CDCN,BCN,chen2021dual} to improve the performance.
Despite good outcomes in the intra-dataset setting, their performances still drop significantly on target domains due to large domain shifts. 
% Face anti-spoofing (FAS) has been studied for decades and could be divided into two stages.
% Initially, the researchers utilize handcrafted features to tackle this problem, such as LBP~\cite{2017Face,maatta2011face,LBP01}, HOG~\cite{2014Context,HoG01} and SIFT~\cite{2016Secure}.
% There are some methods utilizing the information from different domains, such as HSV and YCrCb color spaces~\cite{boulkenafet2015face,boulkenafet2016face}, temporal domains~\cite{siddiqui2016face,bao2009liveness}, and Fourier spectrum~\cite{li2004live}.
% Since the handcrafted features have limited representation capacity, such methods always cannot obtain high performances. With the advent of deep learning, some methods leverage CNN~\cite{DeepBinary00,DeepBinary01,DeepBinary02,zhang2021aurora} to model FAS as a binary classification task. Compared to the handcrafted features, these methods attain higher performance under intra-testing scenarios. However, they still suffer from performance degradation due to the easy overfitting of the training data. Then, other works introduce additional supervision to ease such overfitting, \emph{e.g.,} depth map~\cite{yu2021revisiting}, reflection map~\cite{zhang2021structure} and r-ppg signal~\cite{lin2019face}.
% Based on the above auxiliary information, some methods further adopt different strategies to improve the performance under the intra-dataset setting, such as disentanglement~\cite{disentangle01,STCN} and some custom operators~\cite{CDCN,BCN,chen2021dual}.
% However, the performance of the above methods still drops significantly on an unseen domain. 

\subsubsection{Cross-Domain Face Anti-Spoofing.}
% \noindent \textbf{Cross-Domain Face Anti-Spoofing.} 
To improve the performances under the cross-domain settings, domain generalization (DG)~\cite{li2018domain,li2018learning,zhao2021learning,meng2022attention} is introduced into FAS tasks.
Nevertheless, DG FAS methods~\cite{2019Multi,2020Single,2020Regularized,liu2021dual,liu2021adaptive,chen2021generalizable,zhou2022adaptive} aim to map the samples into a common feature space and lack the specific information of the unseen domains, inevitably resulting in unsatisfactory results.
Considering the availability of the unlabeled target data in real-world applications, several works tackle the above issue based on unsupervised domain adaptation (UDA) methods. Recent studies of UDA FAS mainly rely on pseudo labeling~\cite{quan2021progressive,lv2021combining}, adversarial learning~\cite{2018Unsupervised,DR-UDA,2019Improving,jia2021unified} or minimizing domain discrepancy~\cite{jia2021unified,2018Unsupervised} to narrow the domain shifts.
However, they still suffer from insufficient supervision of the unlabeled target domains, which may cause the negative transfer to the source models. Besides, most works mainly focus on the alignment of high-level semantic features, overlooking the low-level features which are essential to the FAS tasks. In contrast, we aim to address these two issues for UDA FAS.
% However, they still suffer from insufficient supervision of the unlabeled target domains, which may cause the negative transfer to the source models. Besides, most works mainly focus on the alignment of high-level semantic features, overlooking the low-level features which are essential to the FAS tasks. In contrast, we aim to address these two factors for UDA FAS.
% Recent studies in~\cite{quan2021progressive,lv2021combining} utilize the models trained on the source domain to predict the pseudo labels for the unlabeled target data, then transfer the models to the target domains via pseudo labels. The methods in ~\cite{2018Unsupervised,DR-UDA,2019Improving,jia2021unified} adopt the adversarial learning framework or the domain discrepancy loss on high-level semantic features to narrow the domain gap. 
% Although these methods gain better performances on the target domains, they still suffer from the insufficient supervision of the unlabeled target domains, which may cause the negative transfer to the source models. Besides, most works mainly focus on the alignment of high-level semantic features, overlooking the low-level features which are essential to the FAS tasks.
% Thus, the above two factors may affect these methods' performances on the target domains. 

\subsubsection{Unsupervised Domain Adaptation.} Unsupervised domain adaptation (UDA) aims to bridge the domain shifts between the labeled source domain and unlabeled target domain. Recent methods focus on adversarial learning~\cite{ganin2015unsupervised,tzeng2017adversarial,pei2018multi,meng2022slimmable}, self-training~\cite{CBST,CRST,feng2020dmt,xu2021semi}, consistency regularization~\cite{choi2019self,zhou2020uncertainty,zhou2021context,zhou2022domain}, prototypical alignment~\cite{GPA,zhang2021prototypical,jiang2022prototypical}, feature disentanglement~\cite{IID,DISE,zhou2021self} and image translation~\cite{CyCADA,guo2021label,yang2020fda,yang2020phase,liu2021source,isobe2021multi,zhao2022source,PIT,hou2021visualizing,hou2020source}. Despite its gratifying progress, such “model-fitting-to data” paradigm is not practical for FAS task due to plenty of different domains in real-world scenarios. Besides, the discrimination ability of the source model may also be affected during re-training. In contrast, we propose a new yet
practical approach that adapts the target data to the source
model, keeping the source model unchanged. To
the best of our knowledge, this is the first work that reveals
the potential of image translation for UDA FAS.

\begin{figure*}[t!]
	\centering
	\includegraphics[width=1.0\linewidth]{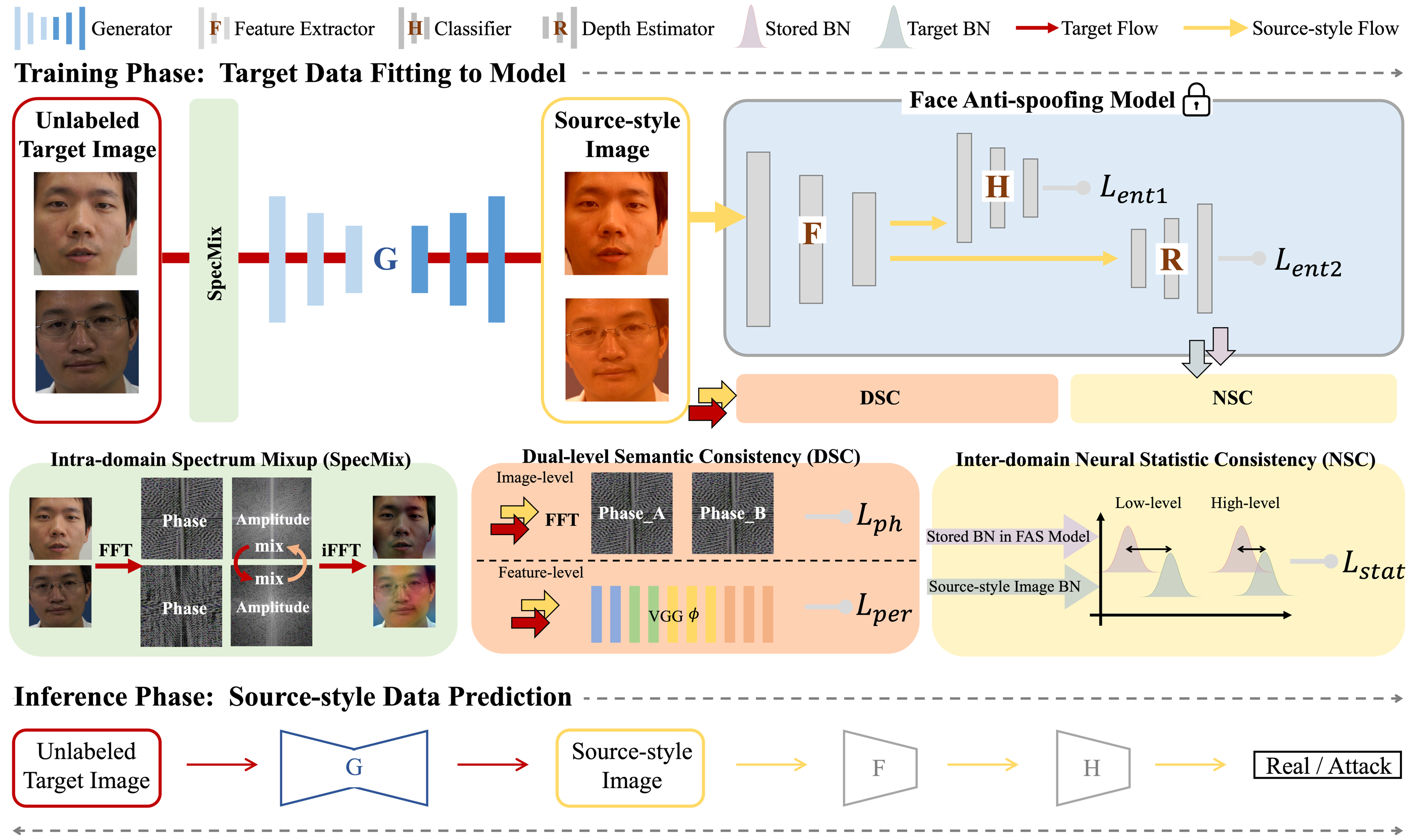}
% 	\vspace{-5mm}
	\caption{Overview of Generative Domain Adaptation framework. 
	The parameters of the source-trained models are fixed during adaptation. 
	Given the unlabeled target data, we only optimize the parameters of the generator $G$. Firstly, we generate diversified target images via \textit{intra-domain spectrum mixup} (SpecMix), thus enhancing the generalization abilities of the generator $G$ in bridging the intra-domain gap. Then, \textit{inter-domain neural statistic consistency} (NSC) fully matches generated feature statistics with the stored ones in high level and low levels, thus mitigating the inter-domain gap. Thus, the translated images can retain the source style. Furthermore, we introduce \textit{dual-level semantic consistency} (DSC) to ensure content-preserving and prevent form semantic distortions
	}
	\label{framework}
%  	\vspace{-4mm}
\end{figure*}

\section{Methodology}
% \subsection{Notations and Overview}
\subsection{Overview}
% In this section, we first introduce some definitions of UDA in FAS tasks and then generally describe our methods.
In UDA FAS, we have access to the labeled source domain, denoted as $D_{s}=\left\{\left(x_{s}, y_{s}\right) \mid x_{s} \subset\right.$ $\left.\mathbb{R}^{H \times W \times 3}, y_{s} \in[0, 1]\right\}$, and the unlabeled target domain, denoted as $D_{t}=\left\{\left(x_{t}\right) \mid x_{t} \subset\right.$ $\left.\mathbb{R}^{H \times W \times 3}\right\}$. 
Regarding that insufficient supervision and neglect of low-level feature alignment in previous UDA FAS approaches lead to inferior performances, we are motivated to perform both the high-level and low-level feature alignment and make the target data fit to the model in a reverse manner.
Our training include two stages: the first phase using the source domains only for training the FAS models, including a feature extractor $F$, a classifier $H$, a depth estimator $R$; the second phase for domain adaptation. During the latter phase, only the image generator $G$ is optimized, and other source models with an ImageNet pre-trained VGG module $\phi$ are fixed during the adaptation. 
% The parameters of $F$, $H$, $R$, and $\phi$ are fixed.
% Our primary goal is to perform the distribution alignment between the labeled source domain $D_{s}$ and the unlabeled target domain $D_{t}$. 

Fig.~\ref{framework} shows the overall GDA framework. We aim to stylize the unlabeled target domain to the source-style domain, making the unlabeled target data fit to the source models, so that the well-trained models do not need to be changed. 
To mitigate the intra-domain gap, input images are firstly diversified in continuous frequency space via intra-domain spectrum mixup (SpecMix) to produce augmented images. 
Then, the generator translates both the original and the diversified target images into the source-style images. To extract the source style information to guide the image translation, we match the generated statistics of the source-style images with those stored source statistics in the pre-trained model via inter-domain neural statistic consistency (NSC), thus bridging the inter-domain gap. Finally, to preserve the target content and prevent semantic distortions during the generation, we propose a dual-level semantic consistency (DSC) on both the feature level and image level. 
% Only the parameters of $G$ are optimized and other source models are fixed during adaptation. 

% \vspace{-3mm}
\subsection{Generative Domain Adaptation}
% \vspace{-3mm}
\subsubsection{Inter-domain Neural Statistic Consistency.}
Batch normalization (BN) \cite{Ioffe2015} normalizes each input feature within a mini-batch in a channel-wise manner so that the output has zero-mean and unit-variance.  Let $B$ and $\{z_i\}_{i=1}^B$ denote the mini-batch size and the input features to the batch normalization, respectively. The key to the BN layer is to compute the batch-wise statistics, \emph{e.g.,} means $\mu$ and variances $\sigma^2$ of the features within the mini-batch:
\begin{eqnarray}
\mu \leftarrow \frac{1}{B} \sum_{i=1}^{B} x_{i}, \ \sigma^{2} \leftarrow \frac{1}{B} \sum_{i=1}^{B}\left(x_{i}-\mu\right)^{2},
\end{eqnarray}
% Then, it normalizes the input features by using the computed BN statistics.
In the first phase of training FAS models, the source statistics $\bar{\mu}^{n+1}_s $ and $\bar{\sigma}^{n+1}_s$ at step $n+1$ are exponential moving average of that at step $n$ with a ratio $\alpha$:
\begin{equation}
\begin{aligned}
\bar{\mu}^{n+1}_s &=(1-\alpha) \bar{\mu}^{n}_s+\alpha \mu^{n}_s \\
\left(\bar{\sigma}^{n+1}_s\right)^{2} &=(1-\alpha)\left(\bar{\sigma}^{n}_s\right)^{2}+\alpha\left(\sigma^{n}_s\right)^{2}
\end{aligned}
\end{equation}
We observe that such neural statistics~\cite{santurkar2018does,Ioffe2015} of the source features
stored in the well-trained FAS models provide sufficient supervisions for both the low-level and high-level features, which can represent domain-specific styles and could be fully used to aid the distribution alignment in UDA. However, the previous methods only use 
the output features of high-level layers
for distribution alignment, and are 
unable to make full use of rich and discriminative liveness cues in low-level features, which is vital to FAS tasks.
Thus, given those stored BN statistics, we can easily estimate the source-style distribution $D_{\tilde{s}}$, where $D_{\tilde{s}}=\left\{\left(x_{\tilde{s}}\right) \mid x_{\tilde{s}}=G\left(x_{t}\right) \subset \mathbb{R}^{H \times W \times 3}\right\}$.
% \begin{equation}
%     D_{\tilde{s}}=\left\{\left(x_{\tilde{s}}\right) \mid x_{\tilde{s}}=G\left(x_{t}\right) \subset \mathbb{R}^{H \times W \times 3}\right\}
% \end{equation}

Inspired by data-free knowledge distillation~\cite{yin2020dreaming}, we propose an \textit{inter-domain neural statistic consistency loss} $\mathcal{L}_{\text {stat }}$ to match the feature statistics between the running mean $\bar\mu_{\tilde{s}}$, running variances $\bar{\sigma}_{\tilde{s}}$ of the source-style data $D_{\tilde{s}}$ and the stored statistics $\bar{\mu_s}$, $\bar{\sigma}_s$ of source models $M$, thus bridging the inter-domain gap:
\begin{equation}
\mathcal{L}_{\text {stat }}=\frac{1}{L} \sum_{l=1}^{L} \left\{ \left\|\bar{\mu}_{\tilde{s}}^{l}-\bar{\mu_{\mathrm{s}}}^{l}\right\|_{2}+\left\|\bar{\sigma}_{\tilde{s}}^{l}-\bar{\sigma}_{\mathrm{s}}^{l}\right\|_{2} \right\}
\end{equation}
where $l \in \{1, 2, ..., L \} $ denotes the layer $l$ in the source-trained models, including the feature extractor $F$, classifier $H$, and depth estimator $R$. Guided by loss $\mathcal{L}_{\text {stat }}$, we could approximate the source-style domain that has the similar style as the source domain.  Different from \cite{yin2020dreaming} that generates image contents from an input random noise, our NSC uses BN statistics alignment as one constraint to stylize the input images without changing contents.

\subsubsection{Dual-level Semantic Consistency.}
To preserve the semantic contents during the image translation, we propose a \textit{dual-level semantic consistency} on both feature level and image level to constrain the contents.

On the feature level, given the generated source-style image $x_{\tilde{s}}$ and the original target image $x_t$ as inputs, a perceptual loss $\mathcal{L}_{\text {per}}$ is imposed onto the latent features of the ImageNet pre-trained VGG module $\phi$, thus narrowing the perceptual differences between them:
% \begin{equation}
% \mathcal{L}_{\text {per}}^{\phi, j}(x_{\tilde{s}}, x_s)=\frac{1}{C_{j} H_{j} W_{j}}\left\|\phi_{j}(x_{\tilde{s}})-\phi_{j}(x_t)\right\|_{2}^{2}
% \end{equation}
\begin{equation}
\mathcal{L}_{\text {per}}^{\phi}(x_{\tilde{s}}, x_s)=\frac{1}{C H W}\left\|\phi(x_{\tilde{s}})-\phi(x_t)\right\|_{2}^{2}
\end{equation}
However, merely using this perceptual loss in the spatial space is not powerful enough to ensure semantic consistency. This is mainly because the latent features are deeply-encoded, and some important semantic cues may be lost. 
% some important spoof cues existing in the low-level features may be lost
Many previous works  \cite{kermisch1970image,piotrowski1982demonstration,oppenheim1981importance,2005Semi,hansen2007structural,yang2020fda,yang2020phase} suggest that the Fourier transformations
from one domain to another only affect the amplitude,
but not the phase of their spectrum, where the phase component retains most of the contents in the original signals, while the amplitude component mainly contains styles.  
And inspired by \cite{yang2020phase}, 
we consider explicitly penalizing the semantic inconsistency  
by ensuring the phase is retained before and after the image translation.
For a single-channel image $x$,  its Fourier transformation $\mathcal{F}(x)$ is formulated:
\begin{equation}
\label{eq:fft}
\mathcal{F}(x)(u, v)=\sum_{h=0}^{H-1} \sum_{w=0}^{W-1} x(h, w) e^{-j 2 \pi\left(\displaystyle{\frac{h}{H}} u+\displaystyle{\frac{w}{W}} v\right)}
\end{equation} 

As such, we enforce phase consistency between the original target image $x_t$ and the source-style image $x_{\tilde{s}}$ by minimizing the following loss $ \mathcal{L}_{ph}$:
\begin{equation}
    \mathcal{L}_{ph}(x_{\tilde{s}}, x_t) = - \sum_{j} \dfrac{\langle\mathcal{F}(x_t)_j, \mathcal{F}(x_{\tilde{s}})_j\rangle}{\|\mathcal{F}(x_t)_j\|_{2}\cdot\|\mathcal{F}(x_{\tilde{s}})_j\|_{2}}
    \label{eq:phase-constraint}
\end{equation}
where $\langle , \rangle$ is the dot-product, and $\| \cdot \|_{2}$ is the $L_2$ norm. Note that Eq. \eqref{eq:phase-constraint} is the negative cosine distance between the original phases and the generated phases. Therefore, by minimizing  $ \mathcal{L}_{ph}$, we can directly minimize their image-level differences on the Fourier spectrum and keep the phase consistency. 

\subsubsection{Intra-domain Spectrum Mixup.}
Given the unlabeled target data, we observe that the generator cannot perform well due to the lack of consideration of intra-domain domain shifts across different target subsets. If training only on the seen training subsets of the target domain and testing on the unseen testing subsets of the target domain, image qualities of the source-style domain could be less-desired.
As such, we wish to a learn more robust generator $G$ under varying environmental changes, \emph{e.g.,} illumination, color.

Since previous findings \cite{kermisch1970image,piotrowski1982demonstration,oppenheim1981importance,2005Semi,hansen2007structural,yang2020fda,yang2020phase,xu2021fourier} reveal that phase tends to preserve most contents in the Fourier spectrum of signals, while the amplitude mainly contains domain-specific styles, we propose to generate diversified images that retain contents but with new styles in the continuous frequency space. 
Through the FFT algorithm~\cite{nussbaumer1981fast}, we can efficiently compute the Fourier transformation $\mathcal{F}(x_t)$ and its inverse transformation $\mathcal{F}^{-1}(x_t)$ of the target image $x_t \in D_t$ via Eq. \ref{eq:fft}.
The amplitude and phase components are formulated as:
\begin{equation}\label{abs_pha}
\begin{aligned}
\mathcal{A}(x_t)(u, v)&=\left[R^{2}(x_t)(u, v)+I^{2}(x_t)(u, v)\right]^{1 / 2} \\
\mathcal{P}(x_t)(u, v)&=\arctan \left[\frac{I(x_t)(u, v)}{R(x_t)(u, v)}\right],
\end{aligned}
\end{equation}
where $R(x_t)$ and $I(x_t)$ denote the real and imaginary part of $\mathcal{F}(x_t)$, respectively. For RGB images, the Fourier transformation for each channel is computed independently to get the corresponding amplitude and phase components.

Inspired from \cite{zhang2017mixup,xu2021fourier}, we introduce \textit{intra-domain spectrum mixup} (SpecMix) by linearly interpolating between the amplitude spectrums of two arbitrary images $x_{t}^{k}$, $x_{t}^{k'}$ from the same  unlabeled target domain $D_t$:
\begin{equation}
\label{eq:specmix}
\hat{\mathcal{A}}(x_{t}^{k}) = (1-\lambda)\mathcal{A}(x_{t}^{k}) + \lambda \mathcal{A}(x_{t}^{k'}),
\end{equation} 
where $\lambda \sim U(0, \eta)$, and the hyper-parameter $\eta$ controls the strength of the augmentation. The mixed amplitude spectrum is then combined with the original phase spectrum to reconstruct a new Fourier representation:
\begin{equation}
\mathcal{F}(\hat{x}_{t}^{k})(u, v)=\hat{\mathcal{A}}(x_{t}^{k})(u, v) * e^{-j *  \mathcal{P}(x_{t}^{k})(u, v)},
\end{equation}
which is then fed to inverse Fourier transformation to generate the interpolated image: $\hat{x}_{t}^{k}=\mathcal{F}^{-1}[\mathcal{F}(\hat{x}_{t}^{k})(u, v)]$.

This proposed \textit{intra-domain spectrum mixup} is illustrated in Fig \ref{framework}. By conducting the aforementioned steps, we could generate unseen target samples with new style and the original content in continuous frequency space. Thus, by feeding forward those diversified images to the generator $G$, the generalization abilities across different subsets within the target domain could be further enhanced.

% Then, we feed the diversified target images to the generator for adaptation, thus enhancing the generator's generalization abilities to capture the intra-domain gap. 

\subsection{Overall Objective and Optimization}
\label{sec3.3}
\noindent \textbf{Entropy loss.}
Minimizing the Shannon entropy of the label probability distribution has been proved to be effective in normal UDA task~\cite{morerio2017minimal,vu2019advent,wu2021entropy,prabhu2021sentry,2020Fully}.
In this paper, we compute entropy loss via the classifier and depth estimator, respectively. The total entropy loss are penelized with $\mathcal{L}_{ent} = \mathcal{L}_{ent1} + \mathcal{L}_{ent2}$. 
 \begin{equation}
 \label{eq:robust-entropy}
\begin{aligned}
 & \mathcal{L}_{ent1} = \sum_{c=1}^{C} -\langle p_c(x_{\tilde{s}}) \cdot \log( p_c(x_{\tilde{s}}) )\rangle  \\
 & \mathcal{L}_{ent2} = \sum_{c=1}^{C} \sum_{h=1}^{H} \sum_{w=1}^{W} -\langle r_c(x_{\tilde{s}})(h, w) \cdot \log( r_c(x_{\tilde{s}}(h,w)) )\rangle  
%  & \mathcal{L}_{ent} = \lambda_{ent1} \mathcal{L}_{ent1} + \lambda_{ent2} \mathcal{L}_{ent2}
\end{aligned}
\end{equation}

\noindent \textbf{Total loss.} During the adaptation procedure, the parameters of the source model $F$, $H$ $R$, and the VGG module $\phi$ are fixed, and we only optimize the parameters of
the generator $G$. The total loss $L_{total}$ is the weighted sum
of the aforementioned loss functions:
\begin{align}
\label{eq:total}
  \mathcal{L}_{total} = \mathcal{L}_{stat} + \mathcal{L}_{per}  + \lambda_{ent} \mathcal{L}_{ent} +  \lambda_{ph} \mathcal{L}_{ph},
\end{align}
where $\lambda_{ent}$, $\lambda_{ph}$, are the weighting coefficients for the loss $\mathcal{L}_{ent}$, $\mathcal{L}_{ph}$ respectively.  
\section{Experiments}
In this section, we first describe the experimental setup in Section~\ref{sec:4.1}, including the benchmark datasets and the implementation details. Then, in Section~\ref{sec:4.2}, we demonstrate the effectiveness of our proposed method compared to the state-of-the-art approaches and related works on multi-source scenarios and single-source scenarios. Next, in Section~\ref{sec:4.3},  we conduct ablation studies to investigate the role of each component in the method. Finally, we provide more visualization and analysis in Section~\ref{sec:4.4} to reveal the insights of the proposed method.

\subsection{Experimental Setup}
\label{sec:4.1}

\noindent \textbf{Datasets.} We use four public datasets that are widely-used in FAS research to evaluate the effectiveness of our method: OULU-NPU \cite{2017OULU} (denoted as O), CASIA-MFSD \cite{Zhang2012A} (denoted as C),
Idiap Replay-Attack \cite{2012Replay} (denoted as I), and MSU-MFSD \cite{2015Face} (denoted as M). 
Strictly following the same protocols as previous UDA FAS methods~\cite{wang2021vlad,wang2021self,jia2021unified,DR-UDA,lv2021combining}, we use source domains to train the source model, the training set of the target domain for adaptation, and the testing set of the target domain for inference. 
% In the multi-source scenarios, we have four tasks in total: O\&C\&I to M, O\&M\&I to C, O\&C\&M to I, and I\&C\&M to O. 
Half Total Error Rate (HTER) and Area Under Curve (AUC) are used as the evaluation metrics~\cite{2019Multi}.

\noindent \textbf{Implementation Details.} Our method is implemented via PyTorch on $24G$ NVIDIA $3090$Ti GPU. We use the same backbone as existing works~\cite{liu2021dual,liu2021adaptive}. Note that we do not use any domain generalization techniques but just a binary classification loss and a depth loss during the first stage. We extract RGB channels of images, thus the input size is $256\times256\times3$. In the second stage, the coefficients $\lambda_{ph}$ and $\lambda_{ent}$ are set to $0.01$ and $0.01$ respectively. The generator $G$~\cite{zhu2017unpaired} is trained with the Adam optimizer with a learning rate of 1e-4.

\begin{figure*}[t!]
\centering
\includegraphics[width=1.0\textwidth]{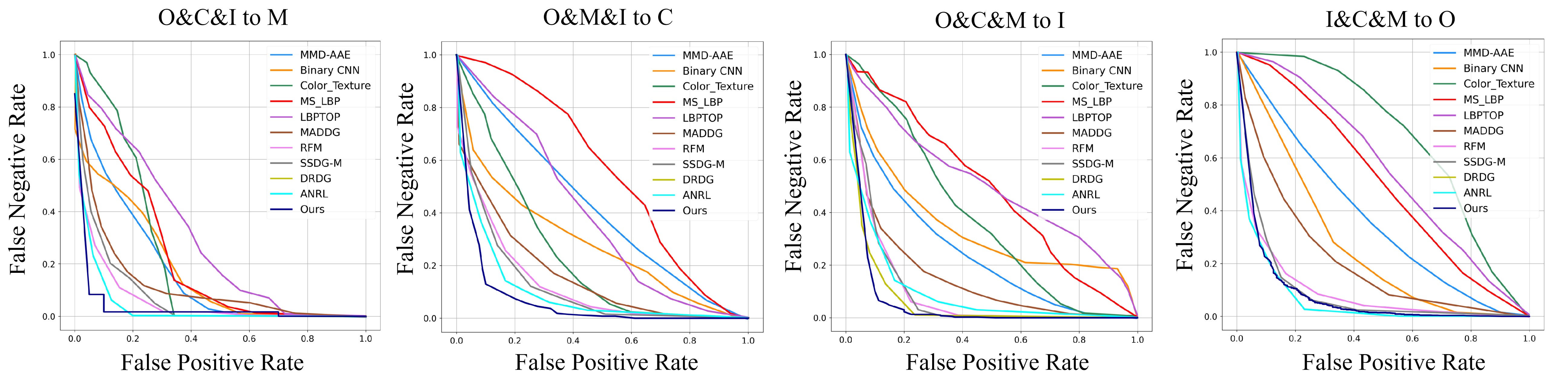}
\caption{ROC curves compared to the state-of-the-art FAS approaches}
\label{fig:roc}
\end{figure*}

\begin{table*}[t!]
\centering
\begin{center}
\caption{Comparisons to the-state-of-art FAS methods on four testing domains}
% \vspace{-2mm}
\label{tab:DG_SOTA}
\resizebox{\textwidth}{!}{%
\begin{tabular}{c | c c | c c | c c | c c }
\hline
\multirow{2}{*}{\textbf{Methods}} &
\multicolumn{2}{c|}{\textbf{O\&C\&I to M}} &
\multicolumn{2}{c|}{\textbf{O\&M\&I to C}} &
\multicolumn{2}{c|}{\textbf{O\&C\&M to I}} &
\multicolumn{2}{c}{\textbf{I\&C\&M to O}}\\

    &HTER(\%) &AUC(\%) &HTER(\%) &AUC(\%) &HTER(\%) &AUC(\%) &HTER(\%) &AUC(\%)\\
\hline
IDA~\cite{2015Face} &$66.6$ &$27.8$ &$55.1$ &$39.0$ &$28.3$ &$78.2$ &$54.2$ &$44.6$\\
LBPTOP~\cite{2014dynamic} &$36.9$ &$70.8$ &$42.6$ &$61.5$ &$49.5$ &$49.5$ &$53.1$ &$44.0$\\
MS\_LBP~\cite{maatta2011face} &$29.7$ &$78.5$ &$54.2$ &$44.9$ &$50.3$ &$51.6$ &$50.2$ &$49.3$\\
ColorTexture~\cite{boulkenafet2016face} &$28.0$ &$78.4$ &$30.5$ &$76.8$ &$40.4$ &$62.7$ &$63.5$ &$32.7$\\
Binary CNN~\cite{2014Learn} &$29.2$ &$82.8$ &$34.8$ &$71.9$ &$34.4$ &$65.8$ &$29.6$ &$77.5$\\
Auxiliary (ALL)~\cite{2018Learning} &- &- &$28.4$ &- &$27.6$ &- &- &-\\
Auxiliary (Depth)~\cite{2018Learning} &$22.7$ &$85.8$ &$33.5$ &$73.1$ &$29.1$ &$71.6$ &$30.1$ &$77.6$\\
\hline
MMD-AAE~\cite{2018Domain} &$27.0$ &$83.1$ &$44.5$ &$58.2$ &$31.5$ &$75.1$ &$40.9$ &$63.0$\\
MADDG~\cite{2019Multi} &$17.6$ &$88.0$ &24.5 &84.5 &$22.1$ &$84.9$ &$27.9$ &$80.0$\\
RFM~\cite{2020Regularized} &13.8 &93.9 &20.2 &88.1 &17.3 &90.4 &16.4 &91.1\\
SSDG-M~\cite{2020Single} &16.7 &90.5 & 23.1 & 85.5 & 18.2 &94.6 & 25.2 & 81.8 \\
DRDG~\cite{liu2021dual} & 12.4 & 95.8 & 19.1 & 88.8 & 15.6 & 91.8 &15.6 & 91.8 \\
ANRL~\cite{liu2021adaptive} & 10.8 & 96.8 & 17.9 & 89.3 & 16.0 & 91.0 & 15.7 & 91.9 \\
\hline
SDA-FAS~\cite{wang2021self} &15.4 &91.8 &24.5 &84.4 &15.6 &90.1 &23.1 &84.3\\
DIPE-FAS~\cite{2014dynamic} &18.2 &- &25.5 &- &20.0 &- &17.5 &-\\
VLAD-VSA~\cite{wang2021vlad} &11.4 &96.4 &20.8 &86.3 &12.3 &93.0 &21.2 &86.9 \\
\hline
Ours &\textbf{9.2} &\textbf{98.0} &\textbf{12.2} &\textbf{93.0} &\textbf{10.0} &\textbf{96.0} &\textbf{14.4} &\textbf{92.6}\\
\hline
\end{tabular}}
\end{center}
\end{table*}

\subsection{Comparisons to the State-of-the-art Methods}
\label{sec:4.2} To validate the generalization capability towards the target domain on the FAS task, we perform two experimental settings of UDA FAS, \emph{i.e.,} multi-source domain adaptation and single-source domain adaptation, respectively. 
% We denote these two settings as multi-source free domain adaptation and single-source free domain adaptation.

\noindent \textbf{Comparisons to FAS methods in multi-source adaptation.} As shown in Table~\ref{tab:DG_SOTA} and Fig.~\ref{fig:roc}, our method outperforms all the state-of-the-art FAS methods under four challenging benchmarks, which demonstrates the effectiveness of our method. 
Conventional FAS approaches~\cite{2015Face,2014dynamic,maatta2011face,boulkenafet2016face,2014Learn,2018Learning} do not consider learning the domain-invariant representations across domains and show less-desired performances.
Besides, almost all DG FAS methods~\cite{2018Domain,2019Multi,2020Single,2020Regularized,liu2021adaptive,liu2021dual} lack a clear target direction for generalization, resulting in unsatisfactory performance in the target domain. Our method outperforms all the DG approaches by significant improvements in both HTER and AUC.
A few DA approaches~\cite{wang2021vlad,wang2021self,2014dynamic} conduct the experiments under this multi-source setting, while they all directly fit the model to the target domain with insufficient supervision and neglect the low-level features for adaptation, leading to undesirable outcomes. In contrast, our method is superior to them by a large margin in four challenging benchmarks.

\begin{table*}[t!]
\centering
\setlength{\belowcaptionskip}{-2mm}
\caption{
Comparisons (HTER) to unsupervised domain adaptation methods}
\begin{center}
\label{tab:Single_DA_SOTA}
\resizebox{0.9\textwidth}{!}{%
\begin{tabular}{c |c |c |c |c |c |c |c}
\hline $\ $ Method $\ $& $\ $  C $\longrightarrow$ I $\ $   &    C $\longrightarrow$ M $\ $ & $\ $ I $\longrightarrow$ C $\ $  & $\ $  I $\longrightarrow$ M $\ $  & $\ $  M $\longrightarrow$ C $\ $ & $\ $  M $\longrightarrow$ I $\ $  & $\ $ Average $\ $ \\
\hline
% \hline SourceOnly & $31.4$ & $43.3$ & $56.3$ & $38.8$ & $33.7$ & $20.8$ & $37.4$ \\
ADDA~\cite{tzeng2017adversarial} & $41.8$ & $36.6$  & $49.8$ & $35.1$ & $39.0$ & $35.2$ & $39.6$ \\
DRCN~\cite{ghifary2016deep} & $44.4$ & $27.6$  & $48.9$ & $42.0$ & $28.9$ & $36.8$ & $38.1$ \\
Dup-GAN~\cite{hu2018duplex} & $42.4$ & $33.4$  & $46.5$ & $36.2$ & $27.1$ & $35.4$ & $36.8$ \\
Auxliary~\cite{2018Remote} & $27.6$ & $-$ & $28.4$ & $-$ & $-$ & $-$ & $-$ \\
De-spoof~\cite{2018De-Spoofing} & $28.5$ & $-$ & $41.1$ & $-$ & $-$ & $-$ & $-$ \\
STASN~\cite{Yang_2019_CVPR} & $31.5$ & $-$ & $30.9$ & $-$ & $-$ & $-$ & $-$ \\
Yang \emph{et al.}~\cite{yang2015person} & $49.2$ & $18.1$ & $39.6$ & $36.7$ & $49.6$ & $49.6$ & $40.5$ \\
KSA~\cite{2018Unsupervised} & $39.2$ & $14.3$ & \textbf{26.3} & $33.2$ & \textbf{10.1} & $33.3$ & $26.1$ \\
ADA~\cite{2019Improving} & $17.5$ & $9.3$ & $41.6$ & $30.5$ & $17.7$ & $5.1$ & $20.3$ \\
DIPE-FAS~\cite{lv2021combining} & $18.9$ & $-$ & $30.1$ & $-$ & $-$ & $-$ & $-$ \\
DR-UDA~\cite{DR-UDA} & $15.6$ & $9.0$ & $34.2$ & $29.0$ & $16.8$ & $3.0$ & $17.9$ \\
USDAN-Un~\cite{jia2021unified} & 16.0& 9.2 &30.2 & 25.8 & 13.3 & 3.4 & 16.3 \\
\hline
Ours & \textbf{15.1} & \textbf{5.8} & 29.7 & \textbf{20.8} & 12.2 & \textbf{2.5} & \textbf{14.4} \\
\hline
\end{tabular}
}
\end{center}
\end{table*}

\begin{table*}[t!]
\centering
\caption{Comparison to the source-free domain adaptation methods}
\begin{center}
\resizebox{0.9\textwidth}{!}{%
\begin{tabular}{c | c c | c c | c c | c c }
\hline
\multirow{2}{*}{\textbf{Methods}} &
\multicolumn{2}{c|}{\textbf{O\&C\&I to M}} &
\multicolumn{2}{c|}{\textbf{O\&M\&I to C}} &
\multicolumn{2}{c|}{\textbf{O\&C\&M to I}} &
\multicolumn{2}{c}{\textbf{I\&C\&M to O}}\\
&HTER(\%) &AUC(\%) &HTER(\%) &AUC(\%) &HTER(\%) &AUC(\%) &HTER(\%) &AUC(\%)\\
\hline
AdaBN~\cite{Li2018Adaptive} &$20.5$ &{88.0} &$34.5$ &$72.0$ &$27.7$ &$80.3$ &{28.2} &{80.8}\\
TENT~\cite{2020Fully} &{20.1} &{88.0} &$35.0$ &$71.2$ &$27.2$ &$79.6$ &$28.3$ &$80.7$\\
SDAN~\cite{he2020self} &$17.7$ &$90.0$ &{25.9} &{81.3} &{28.2} &{84.2} &$32.9$ &$75.0$\\
SHOT~\cite{liang2020we} &$15.0$ &$87.6$ &{20.1} &{84.3} &{40.2} &{57.8} &$25.3$ &$78.2$\\
G-SFDA~\cite{yang2021generalized} &{37.5} &{67.8} &{38.9} &{67.2} &{32.6} &{73.6} &{40.4} &{63.7}\\
NRC~\cite{yang2021exploiting} &{15.0} &{87.4} &{47.8} &{52.4} &{22.1} &{82.3} &{26.6} &{78.8}\\
DIPE-FAS~\cite{lv2021combining} &$18.2$ &- &{25.5} &- &{20.0} &- &$17.5$ &-\\
SDA-FAS~\cite{wang2021self} &15.4 &91.8 &24.5 &84.4 &16.4 &92.0 &23.1 &84.3\\
\hline
Ours &\textbf{9.2} &\textbf{98.0} &\textbf{12.2} &\textbf{93.0} &\textbf{10.0} &\textbf{96.0} &\textbf{14.4} &\textbf{92.6}\\
\hline
\end{tabular}
}
\end{center}
\label{tab:DA_SOTA}
\end{table*}
% \noindent \textbf{Comparisons to FAS methods in single-source adaptation.} To make a fair comparison to the normal UDA approaches in the FAS task, we also conduct experiments in single-source scenarios, where source models are pre-trained on the single-source domain. From Table~\ref{tab:Single_DA_SOTA}, it is obvious to find that our proposed approach shows superiority under four of the six adaptation settings and achieves the best average HTER results. In some hard adaptation tasks, \emph{e.g.,} I $\rightarrow$ C, and M $\rightarrow$ C,  we can achieve the competitive results to the state-of-the-art methods. Interestingly, we find that some results in Table~\ref{tab:Single_DA_SOTA} are superior to results of  Table~\ref{tab:DG_SOTA}. We guess the reasons are that when the source domain number is large, it is hard to train a stable $G$ that generate various styles. Instead, when there is only one source domain and especially the source domain is highly similar to the target domain, it is easier to train a good $G$.
\noindent \textbf{Comparisons to FAS methods in single-source adaptation.} To make a fair comparison to the normal UDA approaches in the FAS task, we also conduct experiments in single-source scenarios, where source models are pre-trained on the single-source domain. From Table~\ref{tab:Single_DA_SOTA}, it is obvious to find that our proposed approach shows superiority under four of the six adaptation settings and achieves the best average HTER results. In some hard adaptation tasks, \emph{e.g.,} I $\rightarrow$ C, and M $\rightarrow$ C,  we can achieve the competitive results to the state-of-the-art methods. Interestingly, we find that some results in Table~\ref{tab:Single_DA_SOTA} are superior to results of  Table~\ref{tab:DG_SOTA}. We guess the reason is that when training on multi-source domains, the style distribution is complicated, and it is hard to train a stable generator $G$, leading to inferior performances. Instead, training on only one source domain with simple style distribution is easier to obtain a better generator $G$. 
% Especially when the target domain is highly similar to the source domain, the adaptation performance would be much better.

\begin{figure*}[t!]
\centering
\includegraphics[width=0.9\textwidth]{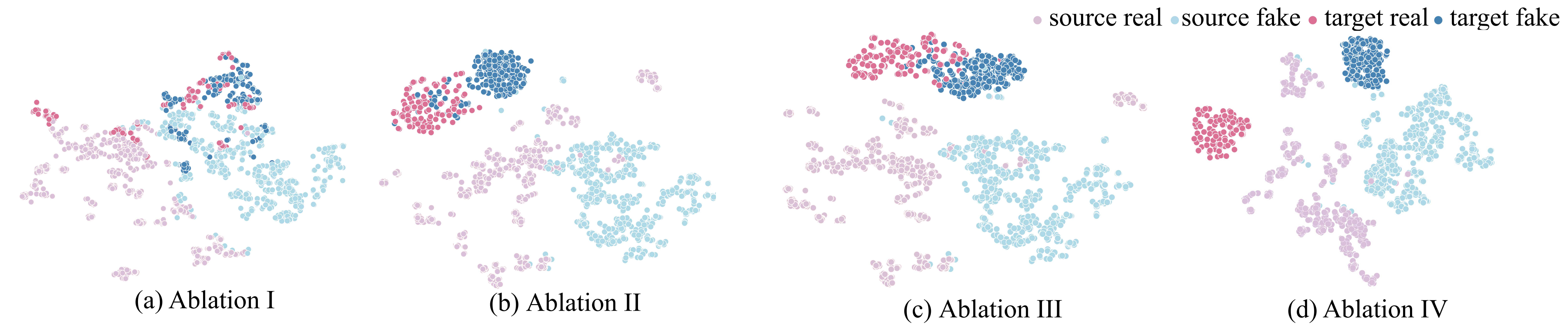}
% \vspace{-7mm}
\caption{The t-SNE visualization of features in different ablation studies}
\label{fig:tsne}
\end{figure*}

\begin{table}[t!]
\caption{Ablation of each component on four benchmarks}
\label{table:ablation}
\centering
\resizebox{0.9\textwidth}{!}{%
\begin{tabular}{c|cccc|cc|cc|cc|cc} 
% \toprule
\hline
\multirow{2}{*}{ID}
&
\multirow{2}{*}{Baseline}
& \multirow{2}{*}{NSC}
& \multirow{2}{*}{DSC} 
& \multirow{2}{*}{SpecMix}
& \multicolumn{2}{c|}{\textbf{O\&C\&I to M}}
& \multicolumn{2}{c|}{\textbf{O\&M\&I to C}}
& \multicolumn{2}{c|}{\textbf{O\&C\&M to I}}
& \multicolumn{2}{c}{\textbf{I\&C\&M to O}} \\
& & & & & HTER & AUC & HTER & AUC & HTER & AUC & HTER & AUC\\
% \midrule
\hline
\uppercase\expandafter{\romannumeral1} & \checkmark  & - & - & - & 29.2 & 77.8 & 32.7 & 76.4 &19.4 & 85.8 &27.1 &80.9  \\
\uppercase\expandafter{\romannumeral2} & \checkmark  & \checkmark  &-  & - & 20.0 & 89.0 & 28.7 & 79.7 & 17.3 & 88.4 &21.1 &84.9 \\
\uppercase\expandafter{\romannumeral3} & \checkmark  & \checkmark  & \checkmark  & - & 14.1 & 92.0 & 14.4 & 90.8  & 13.8 & 91.5 &16.5 &90.7 \\
\uppercase\expandafter{\romannumeral4} & \checkmark  & \checkmark  & \checkmark  & \checkmark &\textbf{9.2} &\textbf{98.0} & \textbf{12.2} & \textbf{93.0} & \textbf{10.0} & \textbf{96.0} &\textbf{14.4} &\textbf{92.6}  \\
% \bottomrule
\hline
\end{tabular}

}
\end{table}

\noindent \textbf{Comparison to the related SFDA methods.} Table \ref{tab:DA_SOTA} presents the comparison results with source-free domain adaptation (SFDA) approaches in four multi-source scenarios. As we can see, if we directly adapt the state-of-the-art SFDA approaches to the FAS task, the performances are less-desired. For example, some unsupervised/self-supervised techniques utilize pseudo labeling~\cite{lv2021combining}, neighborhood clustering~\cite{yang2021exploiting,yang2021generalized}, entropy minimization~\cite{liang2020we,2020Fully} and meta-learning~\cite{wang2021self} to reduce the domain gap between the source pre-trained model and the unlabeled target domain.
The main reasons are two-fold. 1) Almost all of them do not fully utilize the source domain knowledge stored in the pre-trained model, which is not sufficient for feature alignment. 2) They largely neglect the intra-domain gap in the target domain itself, and do not consider learning a more robust domain-invariant representation under varying environmental changes within the target. In contrast, we address these two issues in an explicit manner, and show outstanding improvements on these challenging benchmarks.

\subsection{Ablation Studies}
\label{sec:4.3}
In this section, we perform ablation experiments to investigate the effectiveness of each component, including inter-domain neural statistic consistency (NSC), intra-domain spectrum mixup (SpecMix), dual-level semantic consistency (DSC).

% \begin{table}[t!]
% \caption{Ablation of each component on four benchmarks.}
% \label{table:ablation}
% \centering
% \resizebox{1.0\textwidth}{!}{%
% \begin{tabular}{c|cccc|cc|cc|cc|cc} 
% % \toprule
% \hline
% \multirow{2}{*}{ID}
% &
% \multirow{2}{*}{Baseline}
% & \multirow{2}{*}{NSC}
% & \multirow{2}{*}{DSC} 
% & \multirow{2}{*}{SpecMix}
% & \multicolumn{2}{c|}{\textbf{O\&M\&I to C}}
% & \multicolumn{2}{c|}{\textbf{O\&C\&M to I}}
% & \multicolumn{2}{c|}{\textbf{I\&C\&M to O}}
% & \multicolumn{2}{c}{\textbf{O\&C\&I to M}}\\
% & & & & & HTER & AUC & HTER & AUC & HTER & AUC & HTER & AUC\\
% % \midrule
% \hline
% \uppercase\expandafter{\romannumeral1} & \checkmark  & - & - & - & 32.7 & 76.4 &19.4 & 85.8 &27.1 &80.9 & 29.2 & 77.8 \\
% \uppercase\expandafter{\romannumeral2} & \checkmark  & \checkmark  &-  & -  & 28.7 & 79.7 & 17.3 & 88.4 &21.1 &84.9 & 20.0 & 89.0\\
% \uppercase\expandafter{\romannumeral3} & \checkmark  & \checkmark  & \checkmark  & - & 14.4 & 90.8  & 13.8 & 91.5 &16.5 &90.7 & 14.1 & 92.0\\
% \uppercase\expandafter{\romannumeral4} & \checkmark  & \checkmark  & \checkmark  & \checkmark & \textbf{12.2} & \textbf{93.0} & \textbf{10.0} & \textbf{96.0} &\textbf{14.4} &\textbf{92.6} &\textbf{9.2} &\textbf{98.0} \\
% % \bottomrule
% \hline
% \end{tabular}

% }
% \end{table}

\noindent \textbf{Effectiveness of each component.} Table \ref{table:ablation} shows the ablation studies of each component in four different settings. The baseline means directly feeding forward the target image to the source-trained FAS model for prediction, and the results are with $77.8\%$, $76.4\%$, $85.8\%$, $80.9\%$ AUC, respectively on O\&C\&I to M, O\&M\&I to C, O\&C\&M to I, and I\&C\&M to O, setting. 
By adding NSC, we boost the AUC performances to $89.0\%$, $79.7\%$, $88.4\%$, and $84.9\%$, respectively. Moreover, by adding DSC, we effectively achieve $92.0\%$, $90.8\%$, $91.5\%$, and $90.7\%$, respectively. Finally, our proposed SpecMix effectively increases the performance to $98.0\%$, $93.0\%$, $96.0\%$, $92.6\%$ on four benchmarks, respectively.
These improvements reveal the effectiveness and the complementarities of individual components of our proposed approach. 
% It also reveals that these three components are complementary and together they significantly promote performance.

\noindent \textbf{The t-SNE visualization of features.} To understand how GDA framework aligns the feature representations, we utilize
t-SNE to visualize the feature distributions of both the source and target datasets. As shown in Fig. \ref{fig:tsne}~(a), we observe that the source data can be well discriminated by binary classification, however, the target data can not be well-classified between the real and fake faces without domain adaptation. From Fig. \ref{fig:tsne}~(b), by adding NSC, the classification boundary becomes more clear but there are still some samples misclassified near the decision boundary, the main reason is that it lacks the constraints on the image contents during the generation. As such, by further adding DSC in Fig. \ref{fig:tsne}~(c), the above issue is alleviated to some extent. By bridging both the inter-domain gap and intra-domain gap in Fig. \ref{fig:tsne}~(d), our approach manages to learn a better decision boundary between these two categories, and meanwhile, our target features between different domains become more compact to align.

\setlength{\tabcolsep}{4pt}
\begin{table}[t!]
\begin{center}
\caption{Effect of hyper-parameter $\eta$ of SpecMix}
\label{table:ablations_specmix}
%\ContinuedFloat
% \vspace{2mm}
\subfloat[Effect of SpecMix ($\eta$) on the training \\ of our proposed model on O\&C\&I to M]
{
    \label{table:ablation1_specmix}
    \centering
    % \vspace{2mm}
    \resizebox{0.48\textwidth}{!}
    {%
        \begin{tabular}{c|ccccccc} 
        \hline
        $\eta$ & 0 & 0.1 &0.2 &0.3 &0.4 &0.5 \\
        \hline
        HTER &14.1 &9.2 &10.0 &10.0 &10.0 &10.0  \\
        \hline
        AUC &92.0 &98.0 &97.9 &97.8 &97.3 &97.2  \\
        \hline
        \end{tabular}
    }
}
\subfloat[Effect of SpecMix ($\eta$) on the inference of a well-trained FAS model on Idiap (I)]{
\label{table:ablation2_specmix}
\centering
% \vspace{2mm}
\resizebox{0.46\textwidth}{!}{%
\begin{tabular}{c|ccccccc} 
\hline
$\eta$ & 0 & 0.1&0.2 &0.3&0.4 &0.5\\
\hline
HTER &0 &0 &0 &0 &0 &0.4 \\
\hline
AUC &100 &100 &100 &100 &100 &99.9 \\
\hline
\end{tabular}
}
}
\end{center}
% \vspace{-30pt}
\end{table}

\noindent \textbf{Discussions on SpecMix.} Regarding that SpecMix could generate new styles in continuous frequency space, it it natural to ask several questions: \textit{Will SpecMix change the category during the adaptation? Will mixing amplitude information affect the face liveness? } We conduct several experiments to answer the questions. Note that we use a hyper-parameter $\eta$ to control the strength of augmentation in our SpecMix.  Higher $\eta$ leads to a larger upper bound ratio to mix another image's amplitude from the same batch with that of the current image.
(1). In Table \ref{table:ablations_specmix}~(a), we investigate the effect of $\eta$
 on the training of our proposed GDA framework. During the adaptation, we find that if we set $\eta$=0, which means the intra-domain gaps are neglected, the performance is not perfect. If ranging $\eta$ from 0.1 to 0.5, the performance changes are very slight compared to the best one when setting $\eta=0.1$, but still achieves the state-of-the-art results. (2). As shown in Table  \ref{table:ablations_specmix}~(b), we study the sensitivity of $\eta$ during the inference of a well-trained FAS model on Idiap Replay-Attack dataset. If ranging $\eta$ from 0.1 to 0.4, there are no performance changes, and when $\eta$=0.5, the effect is still slight. From these two aspects, we set $\eta=0.1$ in all experiments, and under such cases, we argue that mixing the amplitude will not affect the face liveness. 

\begin{figure}[t!]
	\centering
    \includegraphics[width=0.9\linewidth]{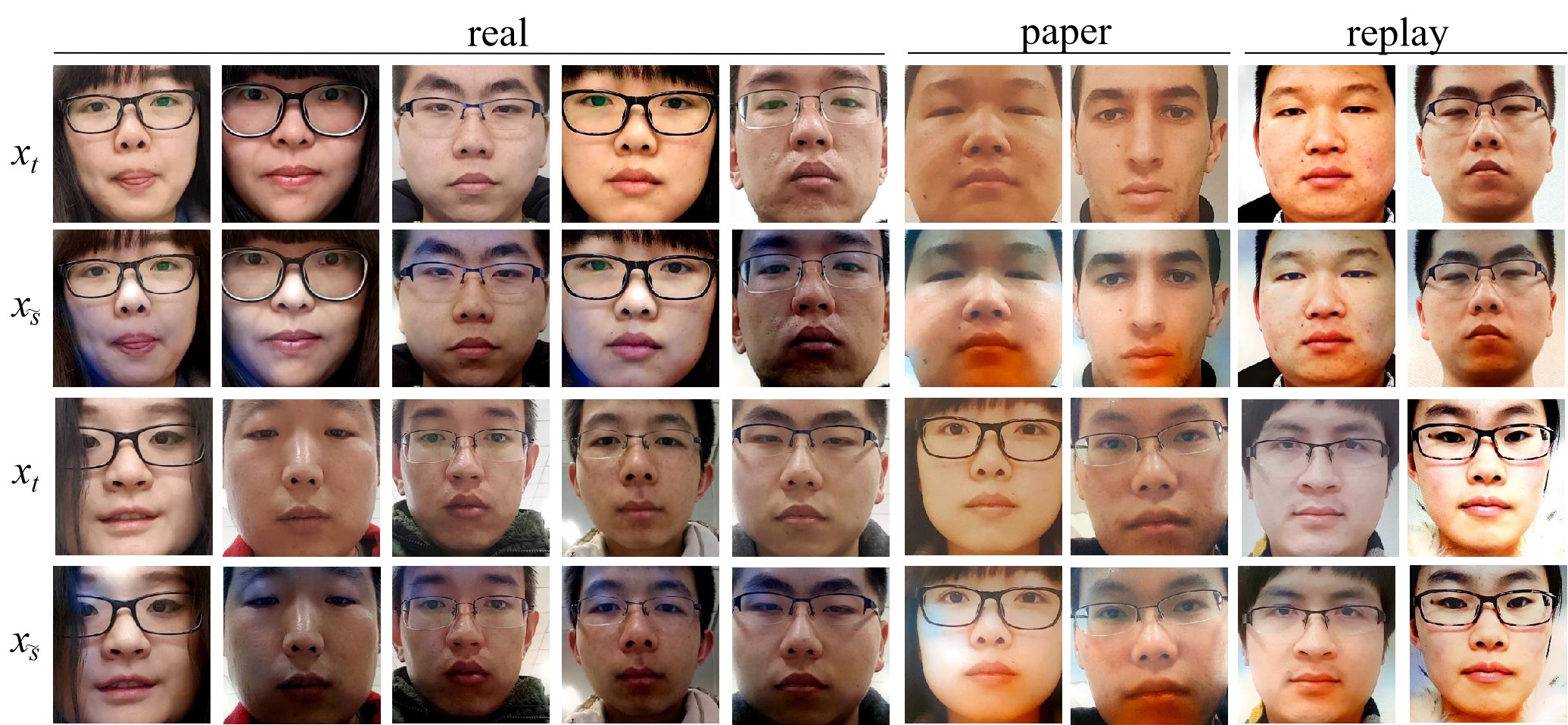}
% 	\vspace{-5mm}
	\caption{Visualization of the target image $x_t$ and the source-style image $x_{\tilde{s}}$}
	\label{fig:vis_gen}
% \vspace{-5mm}
\end{figure}

\subsection{Visualization and Analysis}
\label{sec:4.4}

\begin{figure}[t!]
	\centering
	\includegraphics[width=0.9\linewidth]{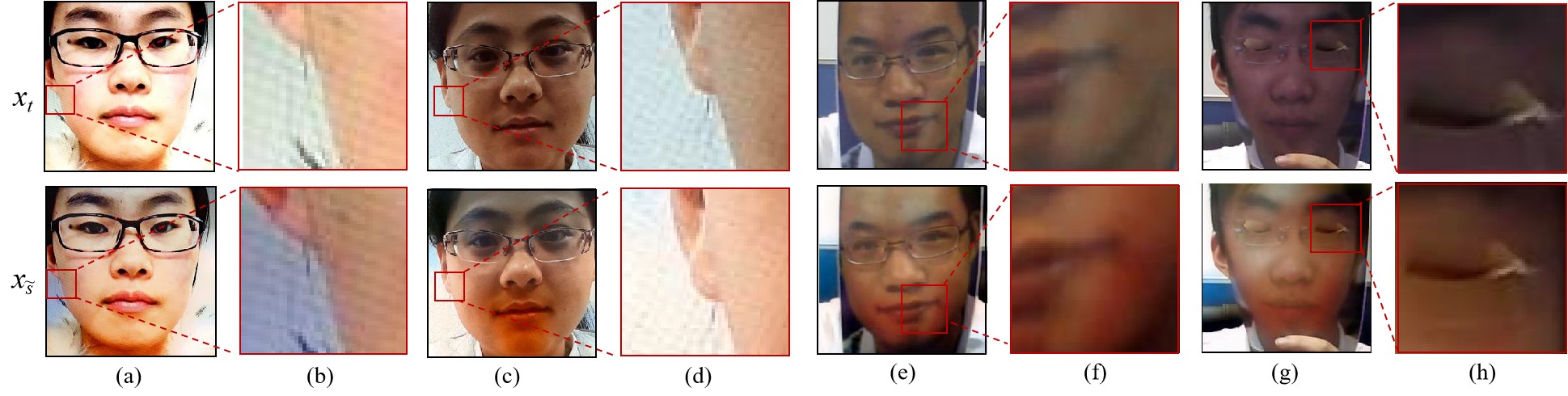}
	\caption{Spoof details in the target images $x_t$ and source-style images $x_{\tilde{s}}$}
	\label{fig:vis_gen_details}
\end{figure}

\noindent \textbf{Visualizations of generated images with source styles.}
To further explore whether the generator succeeds or not in stylizing the target images $x_t$ to a generated images $x_{\tilde{s}}$ that preserves the target content with the source style, we visualize the adapted knowledge in the setting of I\&C\&M to O.  As shown in Fig. \ref{fig:vis_gen}, with the help of our proposed NSC, no matter what kind of faces they are, real faces or fake faces, the style differences between the source domain and the target domain are successfully captured by the generator, which illustrates the effectiveness of our proposed NSC. 
% that succeeds in preserving the source style to recover the pseudo source data during the adaptation. 
Besides, as shown in Fig. \ref{fig:vis_gen_details}, the semantic consistency, especially the spoof details, \emph{e.g.,} moire patterns, paper reflection, are well-maintained between the original target images and the pseudo source images, which demonstrates the effectiveness of our proposed DSC.  
% \textit{More visualizations can be found in the supplemental.}

\begin{figure*}[t!]
\centering
\includegraphics[width=0.9\textwidth]{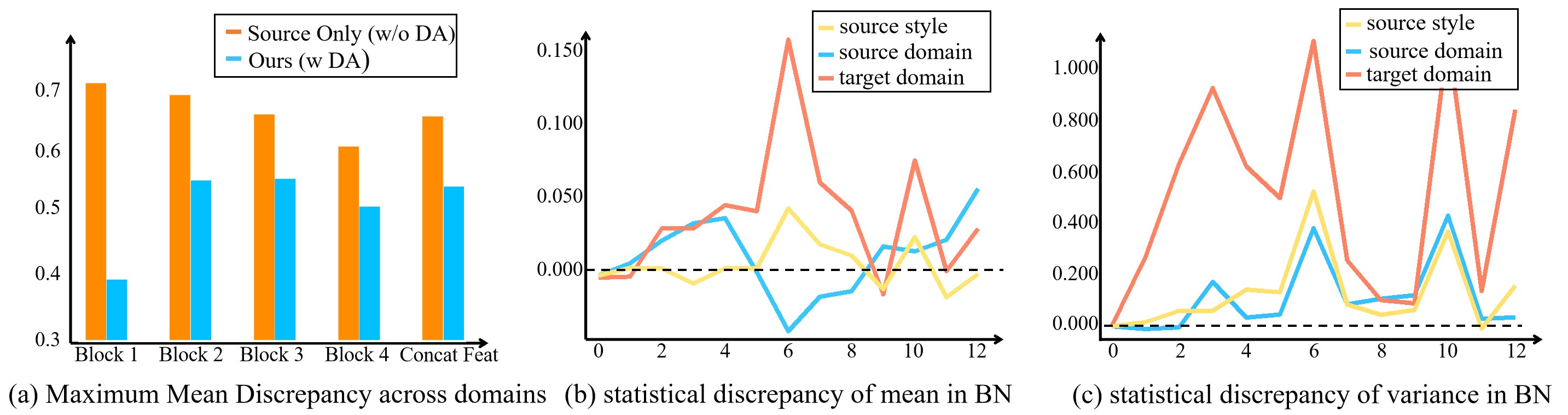}
\caption{Visualization of domain discrepancy (MMD and BN variance) of features}
\label{fig:mmd_bn}
\end{figure*}

\noindent \textbf{Visualizations of cross-domain discrepancy}. 
% To further verify the effectiveness of NSC, we depict the maximum mean discrepancy (MMD) across domains in Fig. \ref{fig:mmd_bn}~(a), and the variation curve of the statistic discrepancy between the current target feature statistics and the source statistics stored in the source-trained model (in Fig. \ref{fig:mmd_bn}~(b) and Fig. \ref{fig:mmd_bn}~(c)). 
As shown in Fig. \ref{fig:mmd_bn}~(a), we measure the maximum mean discrepancy (MMD) across domains. Source Only denotes directly forwarding the testing image to the source model without any domain adaptation techniques. Compared to Source Only, our model effectively reduce the MMD in both the shallow blocks and deep blocks, especially in the shallow blocks, which confirms that our framework successfully translate the target images to the source-style images. In Fig. \ref{fig:mmd_bn}~(b),
we visualize the curve of mean discrepancy and variance discrepancy of each BN layer. As we can see, if directly feeding the target images to the source model in the test phase, the variation of the mean in BN between the source and the target (the red curve) is much larger than our method with domain adaptation methods (yellow curve). Our approach effectively prevents such feature misalignment.
Besides, from Fig. \ref{fig:mmd_bn}~(c), we observe that the variation of variance in BN by feeding the source image to the source model (blue curve) is similar to that with our GDA (yellow curve), which means that our BN statistics effectively align with the source ones. 
% In the left Fig. \ref{fig:mmd_bn}~(b), the misalignment between blue and yellow may result from insufficient source samples, while the variance of yellow cure is smaller than the blue and red curve.
%in contrast to high-level feature representations with large domain gaps and poor transferability, 
\section{Conclusion}
In this work,  we reformulate UDA FAS as a domain stylization problem, aiming to fit the target data onto the well-trained models without changing the models. We propose Generative Domain Adaptation (GDA) framework with several carefully designed components. Firstly, we present an inter-domain neural statistic consistency (NSC) to guide the generator generating the source-style domain. Then, we introduce a dual-level semantic consistency (DSC) to prevent the generation from semantic distortions. Finally, we design an intra-domain spectrum mixup (SpecMix) to reduce the intra-domain gaps. Extensive experiments with analysis demonstrate the effectiveness of our proposed approach.

\noindent \textbf{Acknowledgment:}
This work is supported by National Key Research and Development Program of China (2019YFC1521104), National Natural Science Foundation of China (72192821, 61972157), Shanghai Municipal Science and Technology Major Project  (2021SHZDZX0102), Shanghai Science and Technology Commission (21511101200, 22YF1420300), and Art major project of National Social Science Fund (I8ZD22). 
% Lizhuang Ma and Shouhong Ding are corresponding authors.

\clearpage
\bibliographystyle{splncs04}
\bibliography{egbib}

\begin{thebibliography}{100}
\providecommand{\url}[1]{\texttt{#1}}
\providecommand{\urlprefix}{URL }
\providecommand{\doi}[1]{https://doi.org/#1}

\bibitem{bao2009liveness}
Bao, W., Li, H., Li, N., Jiang, W.: A liveness detection method for face
  recognition based on optical flow field. In: International Conference on
  Image Analysis and Signal Processing. pp. 233--236. IEEE (2009)

\bibitem{boulkenafet2015face}
Boulkenafet, Z., Komulainen, J., Hadid, A.: Face anti-spoofing based on color
  texture analysis. In: IEEE International Conference on Image Processing
  (ICIP). pp. 2636--2640. IEEE (2015)

\bibitem{boulkenafet2016face}
Boulkenafet, Z., Komulainen, J., Hadid, A.: Face spoofing detection using
  colour texture analysis. IEEE Transactions on Information Forensics and
  Security (TIFS)  \textbf{11}(8),  1818--1830 (2016)

\bibitem{2017OULU}
Boulkenafet, Z., Komulainen, J., Li, L., Feng, X., Hadid, A.: Oulu-npu: A
  mobile face presentation attack database with real-world variations. In: 12th
  IEEE international conference on automatic face \& gesture recognition (FG
  2017). pp. 612--618. IEEE (2017)

\bibitem{DISE}
Chang, W.L., Wang, H.P., Peng, W.H., Chiu, W.C.: All about structure: Adapting
  structural information across domains for boosting semantic segmentation. In:
  Proceedings of the IEEE/CVF Conference on Computer Vision and Pattern
  Recognition (CVPR). pp. 1900--1909 (2019)

\bibitem{chen2021dual}
Chen, S., Yao, T., Zhang, K.Y., Chen, Y., Sun, K., Ding, S., Li, J., Huang, F.,
  Ji, R.: A dual-stream framework for 3d mask face presentation attack
  detection. In: Proceedings of the IEEE/CVF International Conference on
  Computer Vision (ICCV). pp. 834--841 (2021)

\bibitem{chen2022daptach}
Chen, Z., Li, B., Wu, S., Xu, J., Ding, S., Zhang, W.: Shape matters:
  Deformable patch attack. In: European conference on computer vision (ECCV)
  (2022)

\bibitem{chen2022ecvit}
Chen, Z., Li, B., Xu, J., Wu, S., Ding, S., Zhang, W.: Towards practical
  certifiable patch defense with vision transformer. In: Proceedings of the
  IEEE/CVF Conference on Computer Vision and Pattern Recognition (CVPR). pp.
  15148--15158 (2022)

\bibitem{chen2021generalizable}
Chen, Z., Yao, T., Sheng, K., Ding, S., Tai, Y., Li, J., Huang, F., Jin, X.:
  Generalizable representation learning for mixture domain face anti-spoofing.
  In: Proceedings of the AAAI Conference on Artificial Intelligence (AAAI).
  vol.~35, pp. 1132--1139 (2021)

\bibitem{2012Replay}
Chingovska, I., Anjos, A., Marcel, S.: On the effectiveness of local binary
  patterns in face anti-spoofing. In: International Conference of Biometrics
  Special Interest Group. pp.~1--7. IEEE (2012)

\bibitem{choi2019self}
Choi, J., Kim, T., Kim, C.: Self-ensembling with gan-based data augmentation
  for domain adaptation in semantic segmentation. In: Proceedings of the
  IEEE/CVF International Conference on Computer Vision (ICCV). pp. 6830--6840
  (2019)

\bibitem{deng2019arcface}
Deng, J., Guo, J., Xue, N., Zafeiriou, S.: Arcface: Additive angular margin
  loss for deep face recognition. In: Proceedings of the IEEE/CVF conference on
  computer vision and pattern recognition (CVPR). pp. 4690--4699 (2019)

\bibitem{DeepBinary00}
Feng, L., Po, L.M., Li, Y., Xu, X., Yuan, F., Cheung, T.C.H., Cheung, K.W.:
  Integration of image quality and motion cues for face anti-spoofing: A neural
  network approach. Journal of Visual Communication and Image Representation
  (JVCIR)  (2016)

\bibitem{feng2020dmt}
Feng, Z., Zhou, Q., Gu, Q., Tan, X., Cheng, G., Lu, X., Shi, J., Ma, L.: Dmt:
  Dynamic mutual training for semi-supervised learning. Pattern Recognition
  (PR) p. 108777 (2022)

\bibitem{2014dynamic}
Freitas~Pereira, T., Komulainen, J., Anjos, A., De~Martino, J., Hadid, A.,
  Pietikäinen, M., Marcel, S.: Face liveness detection using dynamic texture.
  Eurasip Journal on Image and Video Processing  \textbf{2014}(1),  1--15
  (2014)

\bibitem{LBP01}
Freitas~Pereira, T.d., Anjos, A., Martino, J.M.D., Marcel, S.: Lbp- top based
  countermeasure against face spoofing attacks. In: Asian Conference on
  Computer Vision (ACCV). pp. 121--132. Springer (2012)

\bibitem{ganin2015unsupervised}
Ganin, Y., Lempitsky, V.: Unsupervised domain adaptation by backpropagation.
  In: International Conference on Machine Learning (ICML). pp. 1180--1189. PMLR
  (2015)

\bibitem{ghifary2016deep}
Ghifary, M., Kleijn, W.B., Zhang, M., Balduzzi, D., Li, W.: Deep
  reconstruction-classification networks for unsupervised domain adaptation.
  In: European conference on computer vision (ECCV). pp. 597--613. Springer
  (2016)

\bibitem{2005Semi}
Grandvalet, Y., Bengio, Y.: Semi-supervised learning by entropy minimization.
  In: Proceedings of Advances in Neural Information Processing Systems
  (NeurIPS) (2005)

\bibitem{PIT}
Gu, Q., Zhou, Q., Xu, M., Feng, Z., Cheng, G., Lu, X., Shi, J., Ma, L.: Pit:
  Position-invariant transform for cross-fov domain adaptation. In: Proceedings
  of the IEEE/CVF International Conference on Computer Vision (ICCV). pp.
  8761--8770 (2021)

\bibitem{guo2021label}
Guo, S., Zhou, Q., Zhou, Y., Gu, Q., Tang, J., Feng, Z., Ma, L.: Label-free
  regional consistency for image-to-image translation. In: IEEE International
  Conference on Multimedia and Expo (ICME). pp.~1--6. IEEE (2021)

\bibitem{hansen2007structural}
Hansen, B.C., Hess, R.F.: Structural sparseness and spatial phase alignment in
  natural scenes. JOSA A  \textbf{24}(7),  1873--1885 (2007)

\bibitem{he2020self}
He, Y., Carass, A., Zuo, L., Dewey, B.E., Prince, J.L.: Self domain adapted
  network. In: International Conference on Medical Image Computing and
  Computer-Assisted Intervention (MICCAI) (2020)

\bibitem{CyCADA}
Hoffman, J., Tzeng, E., Park, T., Zhu, J.Y., Isola, P., Saenko, K., Efros, A.,
  Darrell, T.: Cycada: Cycle-consistent adversarial domain adaptation. In:
  International conference on machine learning (ICML). pp. 1989--1998. PMLR
  (2018)

\bibitem{hou2020source}
Hou, Y., Zheng, L.: Source free domain adaptation with image translation. arXiv
  preprint arXiv:2008.07514  (2020)

\bibitem{hou2021visualizing}
Hou, Y., Zheng, L.: Visualizing adapted knowledge in domain transfer. In:
  Proceedings of the IEEE/CVF Conference on Computer Vision and Pattern
  Recognition (CVPR). pp. 13824--13833 (2021)

\bibitem{hu2021end}
Hu, C., Zhang, K.Y., Yao, T., Ding, S., Li, J., Huang, F., Ma, L.: An
  end-to-end efficient framework for remote physiological signal sensing. In:
  Proceedings of the IEEE/CVF International Conference on Computer Vision
  (ICCV). pp. 2378--2384 (2021)

\bibitem{hu2018duplex}
Hu, L., Kan, M., Shan, S., Chen, X.: Duplex generative adversarial network for
  unsupervised domain adaptation. In: Proceedings of the IEEE Conference on
  Computer Vision and Pattern Recognition (CVPR). pp. 1498--1507 (2018)

\bibitem{Ioffe2015}
Ioffe, S., Szegedy, C.: Batch normalization: Accelerating deep network training
  by reducing internal covariate shift. In: International Conference on Machine
  Learning (ICML). pp. 448--456 (2015)

\bibitem{isobe2021multi}
Isobe, T., Jia, X., Chen, S., He, J., Shi, Y., Liu, J., Lu, H., Wang, S.:
  Multi-target domain adaptation with collaborative consistency learning. In:
  Proceedings of the IEEE/CVF Conference on Computer Vision and Pattern
  Recognition (CVPR). pp. 8187--8196 (2021)

\bibitem{2020Single}
Jia, Y., Zhang, J., Shan, S., Chen, X.: Single-side domain generalization for
  face anti-spoofing. In: Proceedings of the IEEE Conference on Computer Vision
  and Pattern Recognition (CVPR) (2020)

\bibitem{jia2021unified}
Jia, Y., Zhang, J., Shan, S., Chen, X.: Unified unsupervised and
  semi-supervised domain adaptation network for cross-scenario face
  anti-spoofing. Pattern Recognition (PR)  \textbf{115},  107888 (2021)

\bibitem{jiang2022prototypical}
Jiang, Z., Li, Y., Yang, C., Gao, P., Wang, Y., Tai, Y., Wang, C.: Prototypical
  contrast adaptation for domain adaptive segmentation. In: European Conference
  on Computer Vision (ECCV) (2022)

\bibitem{2018De-Spoofing}
Jourabloo, A., Liu, Y., Liu, X.: Face de-spoofing: Anti-spoofing via noise
  modeling. In: Proceedings of the European conference on computer vision
  (ECCV). pp. 290--306 (2018)

\bibitem{kemelmacher2016megaface}
Kemelmacher-Shlizerman, I., Seitz, S.M., Miller, D., Brossard, E.: The megaface
  benchmark: 1 million faces for recognition at scale. In: Proceedings of the
  IEEE conference on computer vision and pattern recognition (CVPR). pp.
  4873--4882 (2016)

\bibitem{kermisch1970image}
Kermisch, D.: Image reconstruction from phase information only. JOSA
  \textbf{60}(1),  15--17 (1970)

\bibitem{2014Context}
Komulainen, J., Hadid, A., Pietik{\"a}inen, M.: Context based face
  anti-spoofing. In: 2013 IEEE Sixth International Conference on Biometrics:
  Theory, Applications and Systems (BTAS). pp.~1--8. IEEE

\bibitem{Li2018Adaptive}
Li, Yanghao, Wang, Naiyan, Shi, Jianping, Hou, Xiaodi, Liu, Jiaying: Adaptive
  batch normalization for oractical domain adaptation. Pattern Recognition (PR)
   (2018)

\bibitem{li2018learning}
Li, D., Yang, Y., Song, Y.Z., Hospedales, T.: Learning to generalize:
  Meta-learning for domain generalization. In: Proceedings of the AAAI
  conference on artificial intelligence (AAAI). vol.~32 (2018)

\bibitem{2018Unsupervised}
Li, H., Li, W., Cao, H., Wang, S., Huang, F., Kot, A.C.: Unsupervised domain
  adaptation for face anti-spoofing. IEEE Transactions on Information Forensics
  and Security (TIFS)  \textbf{13}(7),  1794--1809 (2018)

\bibitem{li2018domain}
Li, H., Pan, S.J., Wang, S., Kot, A.C.: Domain generalization with adversarial
  feature learning. In: Proceedings of the IEEE conference on computer vision
  and pattern recognition (CVPR). pp. 5400--5409 (2018)

\bibitem{2018Domain}
Li, H., Pan, S.J., Wang, S., Kot, A.C.: Domain generalization with adversarial
  feature learning. In: Proceedings of the IEEE Conference on Computer Vision
  and Pattern Recognition (CVPR). pp. 5400--5409 (2018)

\bibitem{li2004live}
Li, J., Wang, Y., Tan, T., Jain, A.K.: Live face detection based on the
  analysis of fourier spectra. In: Biometric technology for human
  identification. vol.~5404, pp. 296--303. SPIE (2004)

\bibitem{DeepBinary01}
Li, L., Feng, X., Boulkenafet, Z., Xia, Z., Li, M., Hadid, A.: An original face
  anti-spoofing approach using partial convolutional neural network. In:
  International Conference on Image Processing Theory, Tools and Applications
  (IPTA) (2016)

\bibitem{li2021spherical}
Li, S., Xu, J., Xu, X., Shen, P., Li, S., Hooi, B.: Spherical confidence
  learning for face recognition. In: Proceedings of the IEEE/CVF Conference on
  Computer Vision and Pattern Recognition (CVPR). pp. 15629--15637 (2021)

\bibitem{liang2020we}
Liang, J., Hu, D., Feng, J.: Do we really need to access the source data?
  source hypothesis transfer for unsupervised domain adaptation. In:
  International Conference on Machine Learning (ICML). pp. 6028--6039. PMLR
  (2020)

\bibitem{lin2019face}
Lin, B., Li, X., Yu, Z., Zhao, G.: Face liveness detection by rppg features and
  contextual patch-based cnn. In: International Conference on Biometric
  Engineering and Applications (ICBEA) (2019)

\bibitem{liu2021adaptive}
Liu, S., Zhang, K.Y., Yao, T., Bi, M., Ding, S., Li, J., Huang, F., Ma, L.:
  Adaptive normalized representation learning for generalizable face
  anti-spoofing pp. 1469--1477 (2021)

\bibitem{liu2021dual}
Liu, S., Zhang, K.Y., Yao, T., Sheng, K., Ding, S., Tai, Y., Li, J., Xie, Y.,
  Ma, L.: Dual reweighting domain generalization for face presentation attack
  detection. International Joint Conference on Artificial Intelligence (IJCAI)
  (2021)

\bibitem{2018Remote}
Liu, S., Lan, X., Yuen, P.C.: Remote photoplethysmography correspondence
  feature for 3d mask face presentation attack detection. In: Proceedings of
  the European Conference on Computer Vision (ECCV) (2018)

\bibitem{2018Learning}
Liu, Y., Jourabloo, A., Liu, X.: Learning deep models for face anti-spoofing:
  Binary or auxiliary supervision. In: Proceedings of the IEEE conference on
  Computer Vision and Pattern Recognition (CVPR). pp. 389--398 (2018)

\bibitem{STCN}
Liu, Y., Stehouwer, J., Liu, X.: On disentangling spoof trace for generic face
  anti-spoofing. In: European Conference on Computer Vision (ECCV). pp.
  406--422. Springer (2020)

\bibitem{liu2021source}
Liu, Y., Zhang, W., Wang, J.: Source-free domain adaptation for semantic
  segmentation. In: Proceedings of the IEEE/CVF Conference on Computer Vision
  and Pattern Recognition (CVPR). pp. 1215--1224 (2021)

\bibitem{lv2021combining}
Lv, L., Xiang, Y., Li, X., Huang, H., Ruan, R., Xu, X., Fu, Y.: Combining
  dynamic image and prediction ensemble for cross-domain face anti-spoofing.
  In: IEEE International Conference on Acoustics, Speech and Signal Processing
  (ICASSP). pp. 2550--2554 (2021)

\bibitem{maatta2011face}
Maatta, J., Hadid, A., Pietikainen, M.: Face spoofing detection from single
  images using micro-texture analysis. In: Proceedings of the IEEE
  International Joint Conference on Biometrics (IJCB) (2011)

\bibitem{meng2022slimmable}
Meng, R., Chen, W., Yang, S., Song, J., Lin, L., Xie, D., Pu, S., Wang, X.,
  Song, M., Zhuang, Y.: Slimmable domain adaptation. In: Proceedings of the
  IEEE/CVF Conference on Computer Vision and Pattern Recognition (CVPR). pp.
  7141--7150 (2022)

\bibitem{meng2022attention}
Meng, R., Li, X., Chen, W., Yang, S., Song, J., Wang, X., Zhang, L., Song, M.,
  Xie, D., Pu, S.: Attention diversification for domain generalization. In:
  European Conference on Computer Vision (ECCV) (2022)

\bibitem{morerio2017minimal}
Morerio, P., Cavazza, J., Murino, V.: Minimal-entropy correlation alignment for
  unsupervised deep domain adaptation. arXiv preprint arXiv:1711.10288  (2017)

\bibitem{nussbaumer1981fast}
Nussbaumer, H.J.: The fast fourier transform. In: Fast Fourier Transform and
  Convolution Algorithms, pp. 80--111. Springer (1981)

\bibitem{oppenheim1981importance}
Oppenheim, A.V., Lim, J.S.: The importance of phase in signals. Proceedings of
  the IEEE  \textbf{69}(5),  529--541 (1981)

\bibitem{DeepBinary02}
Patel, K., Han, H., Jain, A.K.: Cross-database face antispoofing with robust
  feature representation. In: Chinese Conference on Biometric Recognition. pp.
  611--619. Springer (2016)

\bibitem{2016Secure}
Patel, K., Han, H., Jain, A.K.: Secure face unlock: Spoof detection on
  smartphones. IEEE Transactions on Information Forensics and Security (TIFS)
  \textbf{11}(10),  2268--2283 (2016)

\bibitem{pei2018multi}
Pei, Z., Cao, Z., Long, M., Wang, J.: Multi-adversarial domain adaptation. In:
  Thirty-second AAAI conference on artificial intelligence (AAAI) (2018)

\bibitem{piotrowski1982demonstration}
Piotrowski, L.N., Campbell, F.W.: A demonstration of the visual importance and
  flexibility of spatial-frequency amplitude and phase. Perception
  \textbf{11}(3),  337--346 (1982)

\bibitem{prabhu2021sentry}
Prabhu, V., Khare, S., Kartik, D., Hoffman, J.: Sentry: Selective entropy
  optimization via committee consistency for unsupervised domain adaptation.
  In: Proceedings of the IEEE/CVF International Conference on Computer Vision
  (ICCV). pp. 8558--8567 (2021)

\bibitem{quan2021progressive}
Quan, R., Wu, Y., Yu, X., Yang, Y.: Progressive transfer learning for face
  anti-spoofing. IEEE Transactions on Image Processing (TIP)  \textbf{30},
  3946--3955 (2021)

\bibitem{santurkar2018does}
Santurkar, S., Tsipras, D., Ilyas, A., Madry, A.: How does batch normalization
  help optimization? Advances in neural information processing systems
  \textbf{31} (2018)

\bibitem{2019Multi}
Shao, R., Lan, X., Li, J., Yuen, P.C.: Multi-adversarial discriminative deep
  domain generalization for face presentation attack detection. In: Proceedings
  of the IEEE Conference on Computer Vision and Pattern Recognition (CVPR)
  (2019)

\bibitem{2020Regularized}
Shao, R., Lan, X., Yuen, P.C.: Regularized fine-grained meta face
  anti-spoofing. In: Proceedings of the AAAI Conference on Artificial
  Intelligence (AAAI) (2020)

\bibitem{siddiqui2016face}
Siddiqui, T.A., Bharadwaj, S., Dhamecha, T.I., Agarwal, A., Vatsa, M., Singh,
  R., Ratha, N.: Face anti-spoofing with multifeature videolet aggregation. In:
  2016 23rd International Conference on Pattern Recognition (ICPR). pp.
  1035--1040. IEEE (2016)

\bibitem{taigman2014deepface}
Taigman, Y., Yang, M., Ranzato, M., Wolf, L.: Deepface: Closing the gap to
  human-level performance in face verification. In: Proceedings of the IEEE
  conference on computer vision and pattern recognition (CVPR). pp. 1701--1708
  (2014)

\bibitem{tu2019deep}
Tu, X., Zhang, H., Xie, M., Luo, Y., Zhang, Y., Ma, Z.: Deep transfer across
  domains for face antispoofing. Journal of Electronic Imaging  \textbf{28}(4),
   043001 (2019)

\bibitem{tzeng2017adversarial}
Tzeng, E., Hoffman, J., Saenko, K., Darrell, T.: Adversarial discriminative
  domain adaptation. In: Proceedings of the IEEE conference on computer vision
  and pattern recognition (CVPR). pp. 7167--7176 (2017)

\bibitem{vu2019advent}
Vu, T.H., Jain, H., Bucher, M., Cord, M., P{\'e}rez, P.: Advent: Adversarial
  entropy minimization for domain adaptation in semantic segmentation. In:
  Proceedings of the IEEE/CVF Conference on Computer Vision and Pattern
  Recognition (CVPR). pp. 2517--2526 (2019)

\bibitem{2020Fully}
Wang, D., Shelhamer, E., Liu, S., Olshausen, B., Darrell, T.: Fully test-time
  adaptation by entropy minimization. In: International Conference on Learning
  Representations (ICLR) (2021)

\bibitem{2019Improving}
Wang, G., Han, H., Shan, S., Chen, X.: Improving cross-database face
  presentation attack detection via adversarial domain adaptation. In:
  Proceedings of the IEEE International Conference on Biometrics (ICB) (2019)

\bibitem{DR-UDA}
Wang, G., Han, H., Shan, S., Chen, X.: Unsupervised adversarial domain
  adaptation for cross-domain face presentation attack detection. IEEE
  Transactions on Information Forensics and Security (TIFS)  \textbf{16},
  56--69 (2021)

\bibitem{wang2021self}
Wang, J., Zhang, J., Bian, Y., Cai, Y., Wang, C., Pu, S.: Self-domain
  adaptation for face anti-spoofing. In: Proceedings of the AAAI Conference on
  Artificial Intelligence (AAAI). vol.~35, pp. 2746--2754 (2021)

\bibitem{wang2021vlad}
Wang, J., Zhao, Z., Jin, W., Duan, X., Lei, Z., Huai, B., Wu, Y., He, X.:
  Vlad-vsa: Cross-domain face presentation attack detection with vocabulary
  separation and adaptation. In: Proceedings of the 29th ACM International
  Conference on Multimedia (ACM MM). pp. 1497--1506 (2021)

\bibitem{wang2021facex}
Wang, J., Liu, Y., Hu, Y., Shi, H., Mei, T.: Facex-zoo: A pytorch toolbox for
  face recognition. In: Proceedings of the 29th ACM International Conference on
  Multimedia (ACM MM). pp. 3779--3782 (2021)

\bibitem{2015Face}
Wen, D., Han, H., Jain, A.K.: Face spoof detection with image distortion
  analysis. IEEE Transactions on Information Forensics and Securityn (TIFS)
  \textbf{10}(4),  746--761 (2015)

\bibitem{IID}
Wu, A., Han, Y., Zhu, L., Yang, Y.: Instance-invariant domain adaptive object
  detection via progressive disentanglement. IEEE Transactions on Pattern
  Analysis and Machine Intelligence (TPAMI) pp.~1--1 (2021).
  \doi{10.1109/TPAMI.2021.3060446}

\bibitem{wu2021entropy}
Wu, X., Zhang, S., Zhou, Q., Yang, Z., Zhao, C., Latecki, L.J.: Entropy
  minimization versus diversity maximization for domain adaptation. IEEE
  Transactions on Neural Networks and Learning Systems (TNNLS)  (2021)

\bibitem{xu2021semi}
Xu, H., Liu, F., Zhou, Q., Hao, J., Cao, Z., Feng, Z., Ma, L.: Semi-supervised
  3d object detection via adaptive pseudo-labeling. In: 2021 IEEE International
  Conference on Image Processing (ICIP). pp. 3183--3187. IEEE (2021)

\bibitem{GPA}
Xu, M., Wang, H., Ni, B., Tian, Q., Zhang, W.: Cross-domain detection via
  graph-induced prototype alignment. In: Proceedings of the IEEE/CVF Conference
  on Computer Vision and Pattern Recognition (CVPR). pp. 12355--12364 (2020)

\bibitem{xu2021fourier}
Xu, Q., Zhang, R., Zhang, Y., Wang, Y., Tian, Q.: A fourier-based framework for
  domain generalization. In: Proceedings of the IEEE/CVF Conference on Computer
  Vision and Pattern Recognition (CVPR). pp. 14383--14392 (2021)

\bibitem{2014Learn}
Yang, J., Lei, Z., Li, S.Z.: Learn convolutional neural network for face
  anti-spoofing. In: arXiv preprint arXiv:1408.5601 (2014)

\bibitem{HoG01}
Yang, J., Lei, Z., Liao, S., Li, S.Z.: Face liveness detection with component
  dependent descriptor. In: 2013 International Conference on Biometrics (ICB).
  pp.~1--6. IEEE (2013)

\bibitem{yang2015person}
Yang, J., Lei, Z., Yi, D., Li, S.Z.: Person-specific face antispoofing with
  subject domain adaptation. IEEE Transactions on Information Forensics and
  Security (TIFS)  \textbf{10}(4),  797--809 (2015)

\bibitem{yang2021generalized}
Yang, S., Wang, Y., van~de Weijer, J., Herranz, L., Jui, S.: Generalized
  source-free domain adaptation. In: Proceedings of the IEEE/CVF International
  Conference on Computer Vision (ICCV). pp. 8978--8987 (2021)

\bibitem{yang2021exploiting}
Yang, S., van~de Weijer, J., Herranz, L., Jui, S., et~al.: Exploiting the
  intrinsic neighborhood structure for source-free domain adaptation. In:
  Advances in Neural Information Processing Systems (NeurIPS). vol.~34, pp.
  29393--29405 (2021)

\bibitem{Yang_2019_CVPR}
Yang, X., Luo, W., Bao, L., Gao, Y., Gong, D., Zheng, S., Li, Z., Liu, W.: Face
  anti-spoofing: Model matters, so does data. In: Proceedings of the IEEE/CVF
  Conference on Computer Vision and Pattern Recognition (CVPR). pp. 3507--3516
  (2019)

\bibitem{yang2020phase}
Yang, Y., Lao, D., Sundaramoorthi, G., Soatto, S.: Phase consistent ecological
  domain adaptation. In: Proceedings of the IEEE/CVF Conference on Computer
  Vision and Pattern Recognition (CVPR). pp. 9011--9020 (2020)

\bibitem{yang2020fda}
Yang, Y., Soatto, S.: Fda: Fourier domain adaptation for semantic segmentation.
  In: Proceedings of the IEEE/CVF Conference on Computer Vision and Pattern
  Recognition (CVPR). pp. 4085--4095 (2020)

\bibitem{yin2020dreaming}
Yin, H., Molchanov, P., Alvarez, J.M., Li, Z., Mallya, A., Hoiem, D., Jha,
  N.K., Kautz, J.: Dreaming to distill: Data-free knowledge transfer via
  deepinversion. In: Proceedings of the IEEE/CVF Conference on Computer Vision
  and Pattern Recognition (CVPR). pp. 8715--8724 (2020)

\bibitem{BCN}
Yu, Z., Li, X., Niu, X., Shi, J., Zhao, G.: Face anti-spoofing with human
  material perception. In: European Conference on Computer Vision (ECCV). pp.
  557--575. Springer (2020)

\bibitem{yu2021revisiting}
Yu, Z., Li, X., Shi, J., Xia, Z., Zhao, G.: Revisiting pixel-wise supervision
  for face anti-spoofing. IEEE Transactions on Biometrics, Behavior, and
  Identity Science (TBIOM)  \textbf{3}(3),  285--295 (2021)

\bibitem{CDCN}
Yu, Z., Zhao, C., Wang, Z., Qin, Y., Su, Z., Li, X., Zhou, F., Zhao, G.:
  Searching central difference convolutional networks for face anti-spoofing.
  In: Proceedings of the IEEE/CVF Conference on Computer Vision and Pattern
  Recognition (CVPR). pp. 5295--5305 (2020)

\bibitem{zhang2017mixup}
Zhang, H., Cisse, M., Dauphin, Y.N., Lopez-Paz, D.: mixup: Beyond empirical
  risk minimization. arXiv preprint arXiv:1710.09412  (2017)

\bibitem{zhang2021aurora}
Zhang, J., Tai, Y., Yao, T., Meng, J., Ding, S., Wang, C., Li, J., Huang, F.,
  Ji, R.: Aurora guard: Reliable face anti-spoofing via mobile lighting system.
  arXiv preprint arXiv:2102.00713  (2021)

\bibitem{zhang2021structure}
Zhang, K.Y., Yao, T., Zhang, J., Liu, S., Yin, B., Ding, S., Li, J.: Structure
  destruction and content combination for face anti-spoofing. In: 2021 IEEE
  International Joint Conference on Biometrics (IJCB). pp.~1--6. IEEE (2021)

\bibitem{disentangle01}
Zhang, K.Y., Yao, T., Zhang, J., Tai, Y., Ding, S., Li, J., Huang, F., Song,
  H., Ma, L.: Face anti-spoofing via disentangled representation learning. In:
  European Conference on Computer Vision (ECCV). pp. 641--657. Springer (2020)

\bibitem{zhang2021prototypical}
Zhang, P., Zhang, B., Zhang, T., Chen, D., Wang, Y., Wen, F.: Prototypical
  pseudo label denoising and target structure learning for domain adaptive
  semantic segmentation. In: Proceedings of the IEEE/CVF conference on computer
  vision and pattern recognition (CVPR). pp. 12414--12424 (2021)

\bibitem{Zhang2012A}
Zhang, Z., Yan, J., Liu, S., Lei, Z., Yi, D., Li, S.Z.: A face antispoofing
  database with diverse attacks. In: 2012 5th IAPR international conference on
  Biometrics (ICB). pp. 26--31. IEEE (2012)

\bibitem{zhao2022source}
Zhao, Y., Zhong, Z., Luo, Z., Lee, G.H., Sebe, N.: Source-free open compound
  domain adaptation in semantic segmentation. IEEE Transactions on Circuits and
  Systems for Video Technology (TCSVT)  (2022)

\bibitem{zhao2021learning}
Zhao, Y., Zhong, Z., Yang, F., Luo, Z., Lin, Y., Li, S., Sebe, N.: Learning to
  generalize unseen domains via memory-based multi-source meta-learning for
  person re-identification. In: Proceedings of the IEEE/CVF Conference on
  Computer Vision and Pattern Recognition (CVPR). pp. 6277--6286 (2021)

\bibitem{zhou2019face}
Zhou, F., Gao, C., Chen, F., Li, C., Li, X., Yang, F., Zhao, Y.: Face
  anti-spoofing based on multi-layer domain adaptation. In: 2019 IEEE
  International Conference on Multimedia \& Expo Workshops (ICMEW). pp.
  192--197. IEEE (2019)

\bibitem{zhou2020uncertainty}
Zhou, Q., Feng, Z., Gu, Q., Cheng, G., Lu, X., Shi, J., Ma, L.:
  Uncertainty-aware consistency regularization for cross-domain semantic
  segmentation. Computer Vision and Image Understanding (CVIU) p. 103448 (2022)

\bibitem{zhou2021context}
Zhou, Q., Feng, Z., Gu, Q., Pang, J., Cheng, G., Lu, X., Shi, J., Ma, L.:
  Context-aware mixup for domain adaptive semantic segmentation. arXiv preprint
  arXiv:2108.03557  (2021)

\bibitem{zhou2021self}
Zhou, Q., Gu, Q., Pang, J., Feng, Z., Cheng, G., Lu, X., Shi, J., Ma, L.:
  Self-adversarial disentangling for specific domain adaptation. arXiv preprint
  arXiv:2108.03553  (2021)

\bibitem{zhou2022adaptive}
Zhou, Q., Zhang, K.Y., Yao, T., Yi, R., Ding, S., Ma, L.: Adaptive mixture of
  experts learning for generalizable face anti-spoofing. In: Proceedings of the
  30th ACM International Conference on Multimedia (ACM MM) (2022)

\bibitem{zhou2022domain}
Zhou, Q., Zhuang, C., Lu, X., Ma, L.: Domain adaptive semantic segmentation
  with regional contrastive consistency regularization. In: 2022 IEEE
  International Conference on Multimedia and Expo (ICME). IEEE (2022)

\bibitem{zhu2017unpaired}
Zhu, J.Y., Park, T., Isola, P., Efros, A.A.: Unpaired image-to-image
  translation using cycle-consistent adversarial networks. In: Proceedings of
  the IEEE international conference on computer vision (ICCV). pp. 2223--2232
  (2017)

\bibitem{zhu2022local}
Zhu, W., Wang, C.Y., Tseng, K.L., Lai, S.H., Wang, B.: Local-adaptive face
  recognition via graph-based meta-clustering and regularized adaptation. In:
  Proceedings of the IEEE/CVF Conference on Computer Vision and Pattern
  Recognition (CVPR). pp. 20301--20310 (2022)

\bibitem{CBST}
Zou, Y., Yu, Z., Kumar, B., Wang, J.: Unsupervised domain adaptation for
  semantic segmentation via class-balanced self-training. In: Proceedings of
  the European conference on computer vision (ECCV). pp. 289--305 (2018)

\bibitem{CRST}
Zou, Y., Yu, Z., Liu, X., Kumar, B., Wang, J.: Confidence regularized
  self-training. In: Proceedings of the IEEE/CVF International Conference on
  Computer Vision (ICCV). pp. 5982--5991 (2019)

\end{thebibliography}
\end{document}